\documentclass[runningheads]{llncs}

\usepackage{eccv}

\usepackage{eccvabbrv}

\usepackage{graphicx}
\usepackage{booktabs}
\usepackage{multirow}
\usepackage[inline]{enumitem}
\usepackage{comment}
\usepackage{xcolor}
\usepackage{amssymb}
\usepackage{wrapfig}
\usepackage{subcaption}
\usepackage{tcolorbox}
\usepackage[table]{xcolor} 
\definecolor{highlight}{HTML}{E6F4EA}
\definecolor{purple}{HTML}{6f42f5}

\usepackage[accsupp]{axessibility}  

\usepackage{hyperref}

\usepackage{orcidlink}

\newcommand{\zy}[1]{{\color{blue}[Zhenyu: #1]}}

\begin{document}

\title{DivRL: Disentangled Self-Similarity Rewards for Diverse Subject-Driven Generation} 

\titlerunning{DivRL}

\author{Qian Wang\and
Zhenyu Li\and
Abdelrahman Eldesokey\and
Peter Wonka}

\authorrunning{Q.~Wang et al.}

\institute{KAUST, Saudi Arabia \\
\email{\{first.last\}@kaust.edu.sa}}

\maketitle

\begin{figure}[h]
\centering
\includegraphics[width=\textwidth]{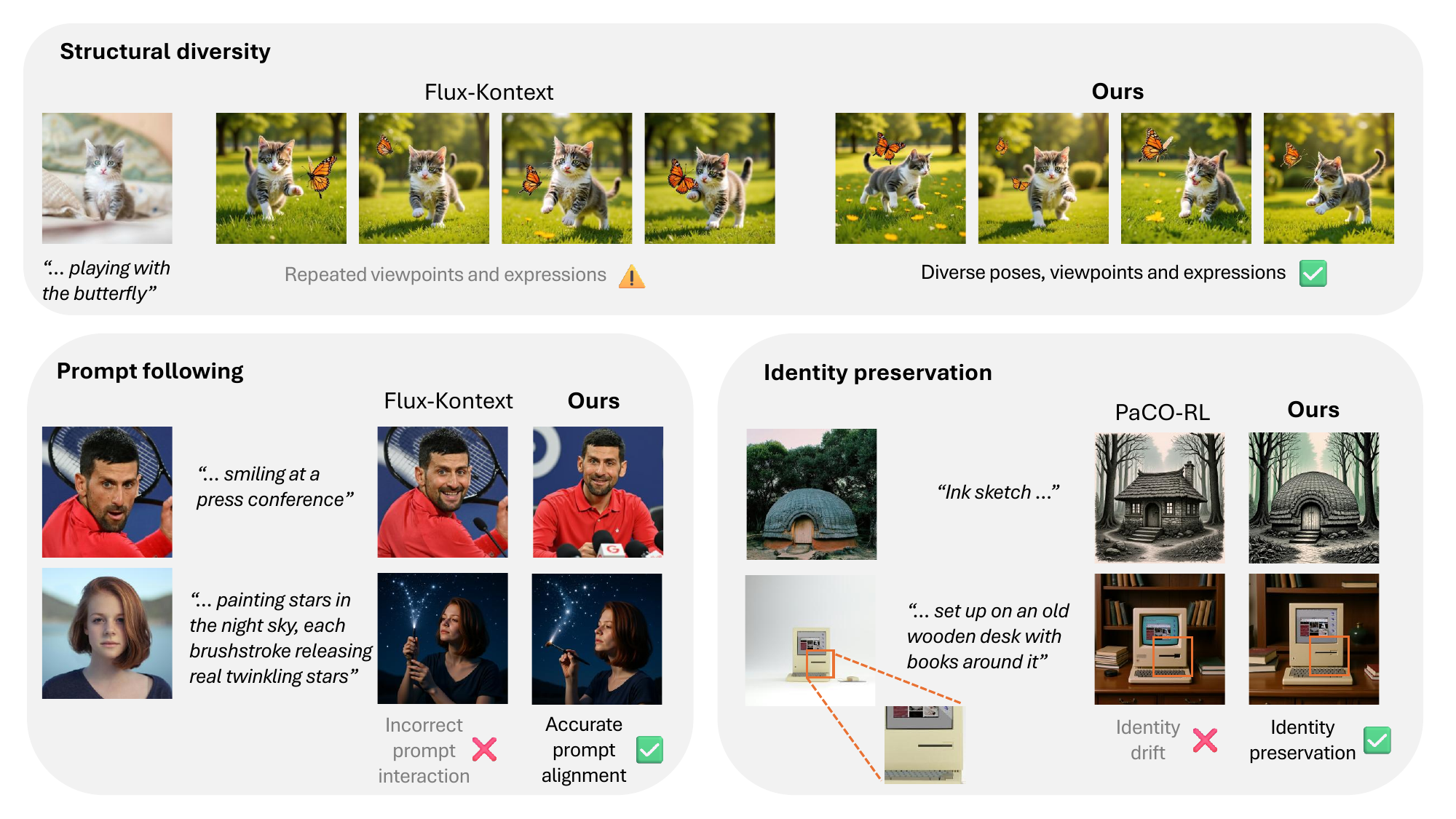}
\caption{Our method achieves a better balance between structural diversity, prompt following, and identity consistency compared with strong baselines.}
\label{fig:teaser}
\end{figure}

\begin{abstract}
  Subject-driven image generation faces an ``Identity-Diversity Paradox'', where strong identity preservation often leads to rigid and low-diversity outputs. 
We propose a post-training framework called \emph{DivRL} that jointly optimizes identity consistency and structural diversity simultaneously by leveraging disentangled visual features from a robust similarity model.
Specifically, we introduce a Negative Self-Similarity Measure (nSSM) to quantify structural diversity, and Visual Semantic Matching (VSM) to evaluate identity consistency. We propose an ``Explore-and-Suppress'' strategy that treats VSM as a gated constraint: the model freely explores structurally diverse configurations, and only samples that violate the identity threshold are penalized via a quadratic hinge loss. This converts identity preservation from a competing objective into a feasibility constraint, allowing nSSM and VSM to improve jointly.
Experiments demonstrate that our method effectively pushes the model to generate both consistent and diverse images and improves structural diversity while maintaining comparable identity consistency through a gated optimization formulation. Code is available at \href{https://github.com/QianWangX/DivRL}{https://github.com/QianWangX/DivRL}.

\end{abstract}

\section{Introduction}
Subject-driven image generation aims to synthesize novel images of a specific subject while preserving its visual identity under diverse contexts, poses, and styles. 
Recent diffusion-based models have demonstrated remarkable fidelity in reproducing reference subjects, enabling applications such as personalized image generation and identity-preserving editing~\cite{labs2025flux1kontext,Xiao2025omnigen,wu2025omnigen2,wu2025qwenimage}. 
However, these models often struggle to balance identity consistency with structural diversity, while adhering to the textual prompt as illustrated in \ref{fig:teaser}. 
Models that strongly enforce identity preservation frequently collapse to near-duplicate images of the reference, producing outputs with similar poses, viewpoints, or expressions. 
Conversely, methods that encourage diversity often drift away from the subject identity. 
This tension between preserving identity and enabling structural variation constitutes what we refer to as the ``Identity–Diversity Paradox''. 

A key reason behind this paradox lies in how identity is represented in a reference image. Identity is defined by a subset of visual attributes, but these attributes are entangled with spatial structure in the reference image. When models attempt to preserve identity, they often also preserve the entangled structural information, which reduces diversity. Conversely, when models attempt to generate images with different structures, they risk modifying identity-defining attributes, leading to identity drift.


Recent work \cite{eldesokey2025mindtheglitch} shows that diffusion backbones contain visual features that are more robust to spatial transformations by disentangling visual and semantic representations. These features provide a reliable signal for evaluating identity consistency even under pose or viewpoint changes. However, optimizing identity consistency using \cite{eldesokey2025mindtheglitch} alone remains insufficient: models can still exploit the reward by reproducing the same spatial configuration as the reference image. 
In other words, identity-based rewards may still encourage structural mimicry by favoring
configurations that closely resemble the reference image. Ideally, we would like the model to preserve identity consistency without sacrificing structural diversity.

To achieve this, we propose a method called \emph{DivRL} for measuring structural diversity through the internal relationships of visual features rather than global feature differences. Specifically, we introduce the \emph{negative Self-Similarity Measure (nSSM)}, which evaluates diversity by comparing the self-similarity matrices of the disentangled visual feature grids extracted from the reference and generated images. Self-similarity matrices capture the intrinsic spatial organization of object parts; therefore, high correlation indicates similar structural layouts, while lower correlation corresponds to structurally diverse contexts. By maximizing nSSM, the model is encouraged to explore structurally different configurations, which often correspond to variations in pose or viewpoint. While we instantiate nSSM using MTG visual features, the formulation applies to any backbone that produces spatially structured feature grids; the identity gate is likewise replaceable by the corresponding backbone's similarity metric.


Reinforcement Learning (RL) allows the model to explore diverse outputs and select structurally novel yet identity-consistent samples. Directly optimizing consistency and diversity objectives together, however, leads to unstable training and reward hacking. 
We therefore introduce an ``Explore-and-Suppress'' optimization strategy built on top of Group Relative Policy Optimization (GRPO). 
The first stage encourages exploration by maximizing structural diversity through nSSM. 
Then, the second stage applies an identity-preserving gate using a visual similarity metric (VSM) \cite{eldesokey2025mindtheglitch}, penalizing samples that deviate excessively from the subject identity. This gated formulation converts the conflict between diversity and identity into a collaborative process: the model first discovers structurally diverse solutions and then undesirable samples are further suppressed during the second-stage optimization.

We evaluate our approach on the DreamBench++ benchmark \cite{peng2024dreambench_plus} using the Flux-Kontext \cite{labs2025flux1kontext} backbone and compare against several subject-driven generation methods. 
Our results show that \emph{DivRL} successfully expands the diversity of generated images while maintaining strong identity consistency, effectively populating the high-utility region where both objectives coexist.

\noindent Our contributions are summarized as follows:
\vspace{-2mm}
\begin{itemize}
\item We introduce the negative Self-Similarity Measure (nSSM), a feature-space metric that quantifies structural diversity via self-similarity correlations of disentangled visual features.
\item We propose an Explore-and-Suppress optimization strategy that decouples diversity exploration from identity preservation using a gated reward formulation.
\item Experiments on DreamBench++ demonstrate that our method improves structural diversity while maintaining competitive identity consistency. We further show that the nSSM formulation and two-stage optimization are backbone-agnostic.
\end{itemize}

\section{Related Work}
\subsection{Subject-driven image generation}
Foundational image diffusion models~\cite{rombach2022sd,esser2024sd3,sauer2024fast} have achieved superior image generation quality. Building on top of the base generative models, a broad line of work tackling the subject-driven image generation fall into three main categories: optimization-Based~\cite{gal2023textualinversion,ruiz2023dreambooth}, adapter-based~\cite{ye2023ip-adapter,li2023blipdiffusion,zhang2023controlnet,chen2024anydoor,shi2024instantbooth,goyal2025shortcut}, and native multimodal models~\cite{labs2025flux1kontext,Xiao2025omnigen,wu2025omnigen2,wu2025qwenimage,liu2025tuna,chen2025blip3}. Optimization-based frameworks typically require a small set of images which represent the reference object, and a dedicated fine-tuning process for every new subject. While enabling faithful personalization, they suffer from lack of high frequency details~\cite{gal2023textualinversion} or catastrophic forgetting~\cite{ruiz2023dreambooth}. Adapter-based methods utilize a pre-trained visual encoder to extract features from a reference image and inject them into a base model. Specifically, \cite{goyal2025shortcut} proposed shortcut-routed adapter training to reduce the copy-paste phenomenon. While offering high efficiency and modularity, they often struggle with complex spatial reasoning and structural flexibility. To overcome these bottlenecks, recently native multimodal models have emerged that utilize a unified diffusion transformer architecture, allowing for more fluid in-context interaction between visual and textual tokens. 
\vspace{-2mm}
\subsection{Reinforcement learning for visual generation}
Inspired by the success of fine-tuning language models with RL~\cite{ouyang2022rlhf,zheng2023secrets,deepseek-math}, emerging works have gained success in aligning image models with the human preferences~\cite{Wallace2024diffusiondpo,zhu2025dspo,na2025boost,black2023ddpo,miao2024training,fan2023dpok,hu2025towards}.
Prominent work Flow-GRPO~\cite{liu2025flowgrpo} integrates GRPO~\cite{deepseek-math} into flow-based models for text-to-image generation. Flow-GRPO converts the deterministic ODE sampling in the flow models into an equivalent SDE, which brings randomness to support the stochastic sampling requirements of the GRPO framework. TempFlow-GRPO~\cite{he2026tempflowgrpo} further proposed noise-aware weighting scheme to prioritize different denoising timesteps during sampling stages. To improve the sampling efficiency of Flow-GRPO, DanceGRPO~\cite{xue2025dancegrpo} selects only specific critical timesteps to update, MixedGRPO~\cite{li2025mixgrpo} applied the gradient updates only within a sliding window of the denoising timestep.

Multiple features-based rewards~\cite{xu2023imagereward,kirstain2023pickscore,wu2023hps,lin2024vqascore} or VLM-based rewards~\cite{luo2025editscore,wu2025editreward,long2026spatialreward}
are designed to measure the image quality from various aspects. To apply RL on the specific subject-driven image generation task, PaCO-RL~\cite{ping2025paco} and Identity-GRPO~\cite{meng2025identitygrpo} proposed pairwise consistency preferences reward modeling, measuring relative ranking of which instance is more consistent than the other given a pair of samples. Unlike these pairwise methods that optimize for relative preference, our framework introduces a gated reward mechanism to explicitly decouple identity preservation from structural mimicry, enabling the model to actively explore regions of the solution space that
simultaneously satisfy identity consistency and structural diversity.

\section{Method}

\begin{figure}[t]
    \centering
    \includegraphics[trim=1cm 8cm 14cm 0cm, width=0.9\linewidth]{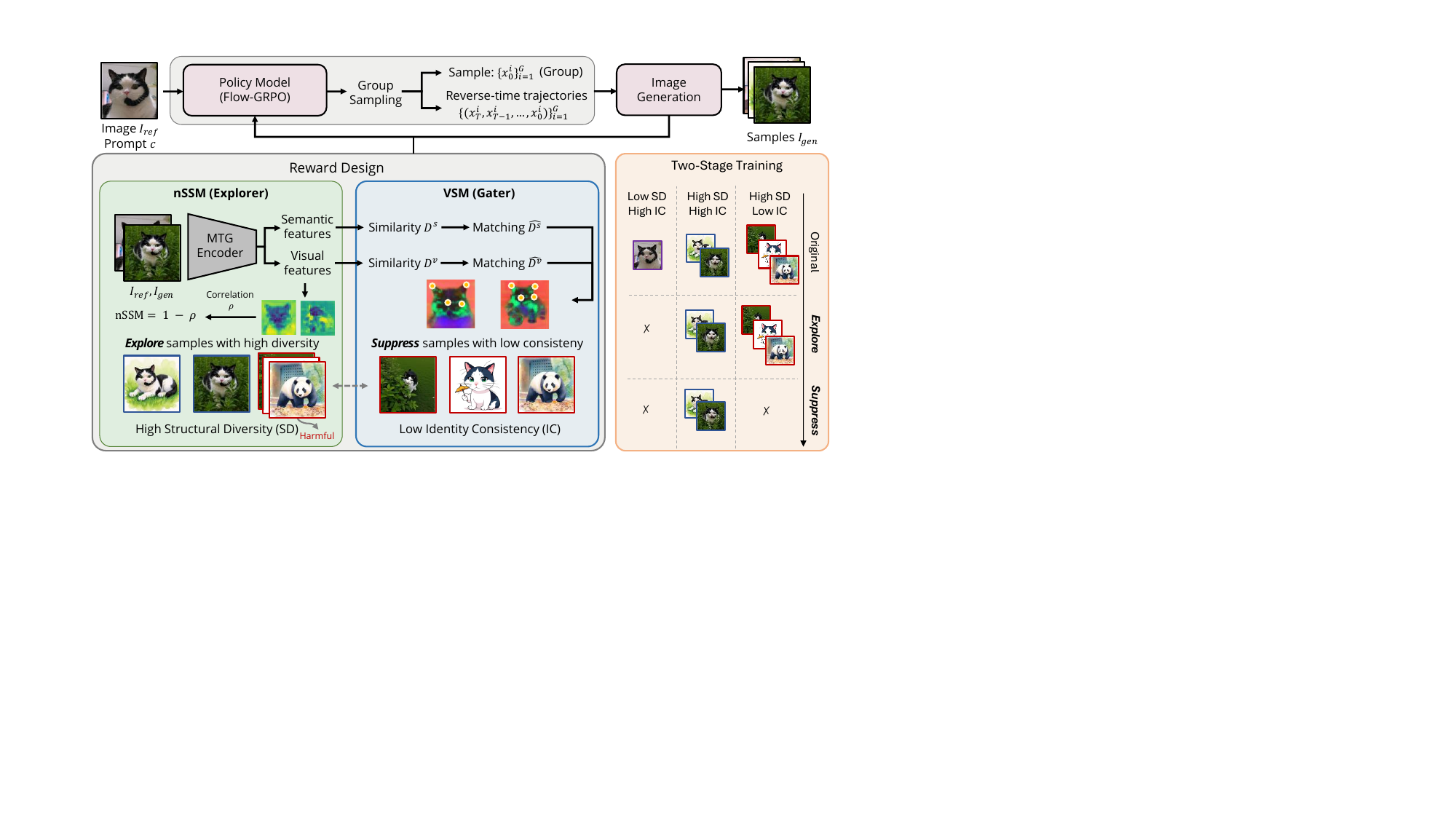}
    \caption{Pipeline of our method DivRL. We design reward models VSM for \textit{identity consistency} (IC) and nSSM for \textit{structural diversity} (SD). A two-stage “Explore-and-Suppress” strategy is used to generate images that are both identity-consistent and structurally diverse. The first stage encourages free exploration for diverse generation, while the second stage employs a gater to filter out low-consistency samples, maintaining both diversity and consistency.}
    \label{fig:method}
\end{figure}

We first introduce the preliminaries about Flow-GRPO. Then, we explain how we design the reward models for identity consistency and structural diversity. 
Afterwards, we explain our ``Explore-and-Suppress'' optimization strategy for aggregating two reward models using a two-stage training process.

\subsection{Preliminaries}
Flow-GRPO optimizes the policy by comparing a group of sampled outputs against their mean reward, which  avoids the need for a separate value function and therefore reduce VRAM consumption. Concretely, for each prompt, multiple responses are sampled and their rewards are normalized within the group to form relative advantages. Given a prompt $c$, the flow model $p_{\theta}$ samples a group of $G$ individual images $\{x_{0}^{i}\}_{i = 1}^G$ and the
corresponding reverse-time trajectories $\{(x_{T}^{i}, x_{T - 1}^{i}, ..., x_{0}^{i})\}_{i = 1}^{G}$. The rewards are calculated on the clean images as $R(x_{0}^i, c)$. The advantage is define as:
\begin{align}
\widehat{A}_{t}^{i} = \frac{R(x_{0}^i, c) - \text{mean}(\{R(x_{0}^i, c)\}_{i = 1}^G)}{\text{std}(\{R(x_{0}^i, c)\}_{i = 1}^G)}.
\end{align}
The policy is then updated to increase the flow objective of samples with above-average rewards while suppressing below-average ones. The objective is to maximize the following objective:
\begin{align}
J_{\theta} &= \frac{1}{G}\sum_{i = 1}^G\frac{1}{T}\sum_{t = 1}^{T} \left(\min(r_t^i(\theta)\widehat{A}_{t}^{i}, \text{clip}(\min(r_t^i(\theta)\widehat{A}_{t}^{i}, \epsilon)) - \beta D_{\text{KL}}(\pi_{\theta}|\pi_{\text{ref}})\right), \notag \\
r &= \frac{\pi_{\theta}}{\pi_{\text{old}}},
\end{align}
where $\pi_{\theta}$ is the current policy and $\pi_{\text{old}}$ is the previous policy, and $D_{\text{KL}}$ is the KL regularization term to keep the current policy not to be too far away from the reference policy $\pi_{\text{ref}}$.

\subsection{Reward Design}
In subject-driven image generation, given a reference image $I_{ref}$ and a text prompt $c$, our goal is to generate both consistent (\ie identity consistent) and diverse (\ie structurally diverse) images $I_{gen}$ that follow the given text prompt.

\subsubsection{Visual Identity Consistency}
In the context of RL for post training, we want to find suitable reward signals for identity consistency and structural diversity.  
Recent work MTG~\cite{eldesokey2025mindtheglitch} extracts fine visual information by disentangling the visual and semantic features from the backbone of the pre-trained diffusion model. These visual features are robust under varying scales, poses, and contexts, making them a valuable metric to evaluate the performance of the identity consistency between generated images and reference images.

Specifically, we forward $I_{ref}$ and $I_{gen}$ to the MTG network to extract semantic features $\mathcal{F}^{s}$ and visual features $\mathcal{F}^{v}$, both with a shape $\mathbb{R}^{48 \times 48 \times c}$, where $c$ is the number of channels. We follow by computing pairwise similarities between individual features to obtain semantic and visual similarity matrices $\mathcal{D}^{s}$ and $\mathcal{D}^{v}$, respectively. The maximum similarity score for each point is taken to compute the best per-point match: $\widehat{\mathcal{D}}^{s} = \max{(\mathcal{D}^{s})}$ and $\widehat{\mathcal{D}}^{v} = \max{(\mathcal{D}^{v})}$. Semantic correspondences are identified by selecting points whose semantic similarity is above threshold $\tau_{s}$, \ie $\widehat{\mathcal{D}}^{s} > \tau_{s}$. Visual consistency is further assessed over semantically consistent regions: 
\begin{align}
\text{VSM}(\tau_{v}) = \frac{1}{\mathcal{J}_{s}}\sum\limits_{j \in \mathcal{J}_{s}}1(\mathcal{D}^{v} > \tau_{v})
\end{align}
where $\mathcal{J}_{s}$ denotes the regions that satisfy $\widehat{\mathcal{D}}^{s} > \tau_{s}$.
A higher VSM generally indicates a higher visual similarity for pixels corresponded to the semantically same object, which imposes a better ID preservation. 
\Cref{fig:method} illustrates this process.

\subsubsection{Structural diversity}
Building on top of the visual features $\mathcal{F}^{v}_{ref}$ for $I_{ref}$ and $\mathcal{F}^{v}_{gen}$ for $I_{gen}$, we further propose negative Self-Similarity Measure (nSSM) for diversity evaluation. 
We first normalize $\mathcal{F}^{v}_{ref}$ and $\mathcal{F}^{v}_{gen}$ across the channel dimension to obtain $\widehat{\mathcal{F}}^{v}_{ref}$ and $\widehat{\mathcal{F}}^{v}_{ref}$ of shape $\mathbb{R}^{48 \times 48}$ each. 
Afterwards, we calculate the SSM matrices for $\widehat{\mathcal{F}}^{v}_{ref}$ and $\widehat{\mathcal{F}}^{v}_{gen}$ to obtain $\mathcal{M}^{v}_{gen}$ and $\mathcal{M}^{v}_{ref}$, respectively. The SSM matrices are calculated as:
\begin{align}
\mathcal{M}^{v}_{ref} = \widehat{\mathcal{F}}^{v}_{ref} (\widehat{\mathcal{F}}^{v}_{ref})^T, \qquad \mathcal{M}^{v}_{gen} = \widehat{\mathcal{F}}^{v}_{gen} (\widehat{\mathcal{F}}^{v}_{gen})^T.
\end{align}
The SSM matrices capture the self-organization of the individual feature patches. We also apply the GT object mask to both $\mathcal{M}^{v}_{ref}$ and $\mathcal{M}^{v}_{ref}$ to obtain $\widehat{\mathcal{M}}^{v}_{ref}$ and $\widehat{\mathcal{M}}^{v}_{gen}$. Afterwards, we compute the Pearson correlation coefficient between $\widehat{\mathcal{M}}^{v}_{ref}$ and $\widehat{\mathcal{M}}^{v}_{gen}$.
\begin{align}
\rho_{(\widehat{\mathcal{M}}^{v}_{ref}, \widehat{\mathcal{M}}^{v}_{gen})} = \frac{\text{cov}(\widehat{\mathcal{M}}^{v}_{ref}, \widehat{\mathcal{M}}^{v}_{gen})}{\sigma_{\widehat{\mathcal{M}}^{v}_{ref}}\sigma_{\widehat{\mathcal{M}}^{v}_{gen}}}.
\end{align}
Intuitively, if the Pearson correlation coefficient is higher, that means the semantic and structural similarity is also higher. As we want to encourage diversity, we take the negative term of $\rho_{(\widehat{\mathcal{M}}^{v}_{ref}, \widehat{\mathcal{M}}^{v}_{gen})}$ and define the final nSSM as 
\begin{align}
\text{nSSM} = 1 - \rho_{(\widehat{\mathcal{M}}^{v}_{ref}, \widehat{\mathcal{M}}^{v}_{gen})}. 
\end{align}
This nSSM reward is computed on dense latent diffusion feature grid, which incorporates fine-grained geometry and can provide rich structural supervision. 


\subsubsection{Stabilizing Convergence}
Recall that the spatial resolution of both $\mathcal{F}^{v}_{ref}$ and $\mathcal{F}^{v}_{gen}$ is $48 \times 48$. 
While this fine-grained resolution can capture rich structure-related details and make it a suitable resolution to evaluate the structural similarity, the model is implicitly encouraged to preserve those dense local correlation patterns during the training stage, leading to high-frequency textural artifacts. 
To solve this issue, we additionally apply a $2 \times 2$ average pooling on the normalized visual features to obtain downsampled features $\widehat{\mathcal{F}}^{v}_{ref}$ and $\widehat{\mathcal{F}}^{v}_{ref}$, respectively, which are then of size $24 \times 24$. The $\widehat{\mathcal{F}}^{v}_{ref}$ is then calculated as:
\begin{align}
\widehat{\mathcal{F}}^{v}_{ref} &= \text{avgpool}_{2\times2}(\text{normalize}(\mathcal{F}^{v}_{ref})),  \notag \\ \widehat{\mathcal{F}}^{v}_{gen} &= \text{avgpool}_{2\times2}(\text{normalize}(\mathcal{F}^{v}_{gen})).
\end{align}

\subsection{Optimization Strategy}
We propose a two-stage optimization strategy to facilitate the RL. The only difference between these two stages lies in the definition of the reward for training. In the first stage, we merely use the nSSM as the reward model:
\begin{align}
R_1 = nSSM,
\end{align}
which effectively encourages the model to explore desirable latent region generating both consistent and diverse samples. However, we empirically observe the emergence of inconsistent samples along the training due to the reward hacking (\ie, generating random samples without any consistency can also increase the reward of nSSM).

In the second stage, we introduce the VSM as a ``gate'' to suppress those noisy samples that  provide inconsistent gradient directions. The reward is formulated as
\begin{align}
R_2 = 
\begin{cases}
\text{nSSM} & \text{VSM} \geq s, \\
\text{nSSM} - \lambda (s - \text{VSM})^2 & \text{VSM} < s,
\end{cases}
\end{align}
where $s$ is a threshold for identity consistency, and $\lambda$ is a weighting term for the penalty of the non-similarity degree. Intuitively, the gated reward transforms identity preservation into a feasibility constraint rather than a competing objective. The model is free to explore diverse structural configurations as long as they remain within the identity-consistent region. The introduction of VSM effectively alleviates the reward hacking issue and further facilitates the optimization of nSSM, reduces the tendency of the model to rely on rigid solutions and explores the high-quality latent space where identity and structural novelty coexist. This avoids the optimization instability observed when diversity and identity are directly combined through linear weighting.

\vspace{-5mm}
\section{Experiments}
\vspace{-2mm}
\subsection{Implementation details}
\subsubsection{Settings} We adopt the vanilla Flow-GRPO framework to optimize the model and use Flux-Kontext~\cite{labs2025flux1kontext} as the backbone for subject-driven image generation. Training is conducted on a 10k subset of the SynCD dataset~\cite{kumari2025syncd}, where each identity is paired with 10 text prompts. For the reward model, we set the semantic and visual thresholds to $\tau_s=0.7$ and $\tau_v=0.7$, respectively. The hinge parameters are set to $\lambda=5$ and $s=0.5$. Training is performed in two stages: 3200 optimization steps for the exploration stage, followed by 3200 steps for the suppression stage. For Flow-GRPO, the group size is set to 21 ,and the number of denoising steps during sampling is 6. The KL regularization coefficient is $\beta=0.1$. During inference, we use 28 denoising steps. Training is conducted on 8 NVIDIA A100 (80GB) GPUs, with each training stage taking 24 hours.
\vspace{-3mm}
\subsubsection{Evaluation}
We evaluate the models from four perspectives: identity consistency, structural diversity, prompt following, and aesthetic quality. For identity consistency, we report the out-of-domain metrics CLIP image cosine similarity and DINO cosine similarity, as well as the in-domain metric VSM. We additionally report the \textit{consistency ratio}, defined as the percentage of generated samples satisfying $\text{VSM}>0.6$ for in-domain evaluation, and structural DINO $\text{sDINO}>0.75$ for out-of-domain evaluation. For structural diversity, we report the in-domain metric MTG-nSSM and out-of-domain metrics including DINO-nSSM, scale-invariant IoU, and LPIPS. DINO-nSSM is computed using the same formulation as MTG-nSSM but with DINO features instead of MTG visual features. We further report \textit{diversity-over-consistency}, which measures diversity among identity-consistent samples. The in-domain version computes the average MTG-nSSM over samples with $\text{VSM}>0.6$, while the out-of-domain version computes the average DINO-nSSM over samples with $\text{sDINO}>0.75$. We will provide the detailed definition of the metrics in the Supplementary Materials. All evaluations are conducted on DreamBench++~\cite{peng2024dreambench_plus}, which contains 150 identities spanning animals, humans, objects, and style transfer scenarios, each paired with 9 text prompts.

\vspace{-3mm}
\subsubsection{Baselines}
We compare our method with Flux-IP-Adapter\footnote{\href{https://huggingface.co/XLabs-AI/flux-ip-adapter}{https://huggingface.co/XLabs-AI/flux-ip-adapter}}, OmniGen~\cite{Xiao2025omnigen}, OmniGen2~\cite{wu2025omnigen2}, UNO~\cite{wu2025uno}, PaCo-RL~\cite{ping2025paco}, and the original Flux-Kontext~\cite{labs2025flux1kontext}. Flux-IP-Adapter injects image conditioning into the Flux backbone through an IP-Adapter module. OmniGen and OmniGen2 are unified multimodal models capable of handling various image-conditioned generation tasks. UNO is a multi-image conditioned multimodal model designed for subject-driven generation. PaCo-RL introduces a pairwise consistency reward model and fine-tunes Flux-Kontext using Flow-GRPO. All methods are evaluated at a resolution of $1024 \times 1024$.

\vspace{-4mm}
\subsection{Quantitative results}

\begin{table}[t]
\centering
\small
\caption{Comparison of different subject-driven generation models across various metrics.}
\label{tab:quant_comp}
\renewcommand{\arraystretch}{1.2} 
\setlength{\tabcolsep}{4pt}      
\resizebox{\textwidth}{!}{
\begin{tabular}{l c | ccc | ccc | c} 
\toprule
& \textbf{Prompt foll.} & \multicolumn{3}{c|}{\textbf{Identity consistency}} & \multicolumn{3}{c|}{\textbf{Diversity}} & \textbf{Aesthetic} \\ 
\textbf{Model} & CLIP-T$\uparrow$ & CLIP-I$\uparrow$ & DINO$\uparrow$ & VSM-0.7$\uparrow$ & DINO-nSSM$\uparrow$ & Scale-inv IOU$\downarrow$ & MTG-nSSM$\uparrow$ & HPS$\uparrow$ \\
\midrule
Flux-IP-Adapter & 0.274 & 0.805 & 0.568 & 0.481 & 0.531 & 0.593 & 0.765 & 0.275 \\
OmniGen         & 0.293 & 0.735 & 0.474 & 0.464 & 0.546 & 0.583 & 0.786 & 0.301 \\
OmniGen2         & 0.301 & 0.802 & 0.559 & 0.460 & 0.496 & 0.638 & 0.727 & 0.311 \\
UNO             & 0.209 & 0.710 & 0.387 & 0.454 & 0.590 & 0.637 & 0.816 & 0.215 \\
\midrule
Flux-Kontext    & \underline{0.283} & 0.781 & 0.594 & 0.605 & 0.433 & 0.694 & 0.659 & \underline{0.294} \\
PaCo-RL         & \textbf{0.287} & 0.766 & 0.570 & 0.576 & 0.445 & \underline{0.683} & 0.668 & \textbf{0.302} \\
\rowcolor{highlight} \textbf{DivRL (VSM only)} & 0.278 & \textbf{0.806} & \textbf{0.670} & \textbf{0.688} & 0.383 & 0.757 & 0.608 & 0.291 \\
\rowcolor{highlight} \textbf{DivRL (nSSM only)} & \textbf{0.287} & 0.756 & 0.520 & 0.562 & \textbf{0.489} & \textbf{0.634} & \textbf{0.704} & 0.293 \\
\rowcolor{highlight} \textbf{DivRL (VSM + nSSM)} & 0.279 & \underline{0.786} & \underline{0.600} & \underline{0.614} & \underline{0.453} & 0.694 & \underline{0.689} & 0.287 \\
\bottomrule
\end{tabular}}
\end{table}

We present a quantitative evaluation of different methods in terms of prompt following, identity consistency, structural diversity ,and aesthetic in Table~\ref{tab:quant_comp}. We categorize the methods into two different categories: Flux-Kontext-based models and others. In addition to our full model, we report two variants trained with a single reward signal: VSM-only and nSSM-only, which illustrate the optimization boundaries of identity consistency and structural diversity. Optimizing with VSM alone significantly improves identity consistency compared to the baselines, but leads to a noticeable drop in diversity metrics. In contrast, optimizing with nSSM alone substantially increases structural diversity but reduces identity consistency. By combining the two rewards through our proposed optimization strategy, our method achieves a better balance between identity consistency and structural diversity. In particular, it improves diversity over the Flux-Kontext baseline while maintaining comparable identity consistency.

Interestingly, even without explicitly optimizing for text alignment, our method achieves strong prompt-following performance, with the nSSM-only variant obtaining the highest text-alignment score. This phenomenon can be explained by the geometric rigidity induced by identity preservation during pretraining. Encouraging structural diversity through nSSM relaxes this rigidity, allowing the model to better adapt the subject to the contextual requirements of the prompt.

\begin{figure}[t!]
\centering
\includegraphics[width=0.7\textwidth]{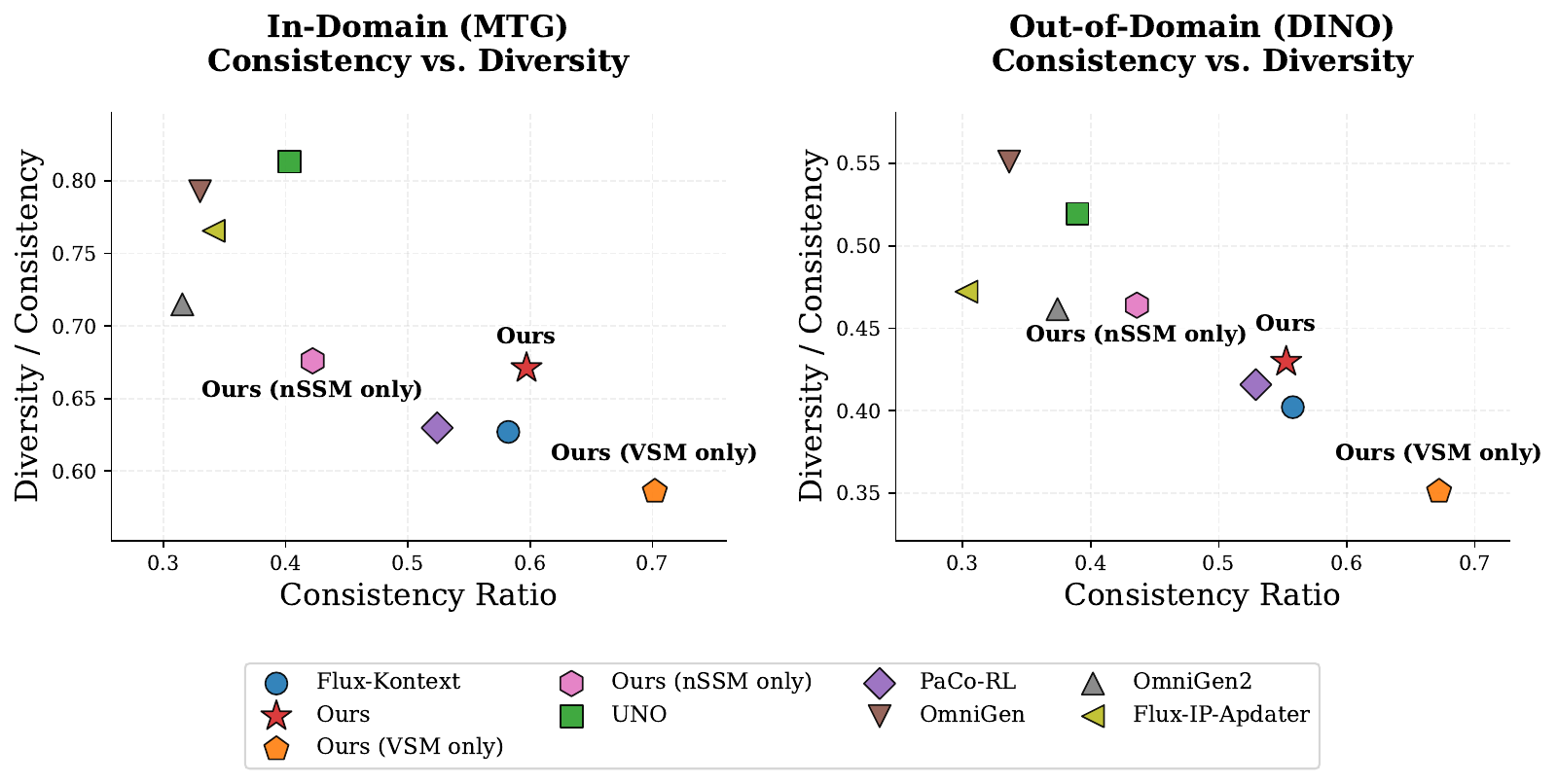}
\caption{Quantitative evaluation of the trade-off between identity consistency and structural diversity. Results are reported using both in-domain metrics (MTG) and out-of-domain metrics (DINO). }
\label{fig:quant_comp}
\end{figure}

We further analyze the trade-off between consistency and diversity in Figure~\ref{fig:quant_comp} using the consistency ratio and diversity-over-consistency metrics. Both in-domain and out-of-domain evaluations show that non-Flux-Kontext-based models achieve high diversity for samples that pass the consistency threshold, but their overall consistency ratio is significantly lower than that of Flux-Kontext-based methods. This suggests that these models struggle to reliably preserve subject identity, which is a fundamental requirement in subject-driven generation. Notably, OmniGen2 demonstrates markedly improved global feature similarity over OmniGen, yet its VSM score and consistency ratio remain low, indicating that it still falls short of reliable fine-grained identity preservation and cannot close the gap with Flux-Kontext-based methods on this dimension.  

Among Flux-Kontext-based methods, the base Flux-Kontext model already achieves strong identity preservation, reflected by its high consistency ratio. PaCo-RL produces slightly more diverse outputs but exhibits a lower consistency ratio compared to the base model. The two variants of our method demonstrate the expected extremes: the nSSM-only model achieves higher diversity but lower consistency, while the VSM-only model achieves the highest consistency ratio but the lowest diversity. Our full method combines the advantages of both variants, maintaining a consistency level comparable to the baselines while achieving improved diversity.

\begin{figure}[ht!]
\centering
\includegraphics[trim=0cm 5.5cm 0cm 0cm, width=\textwidth]{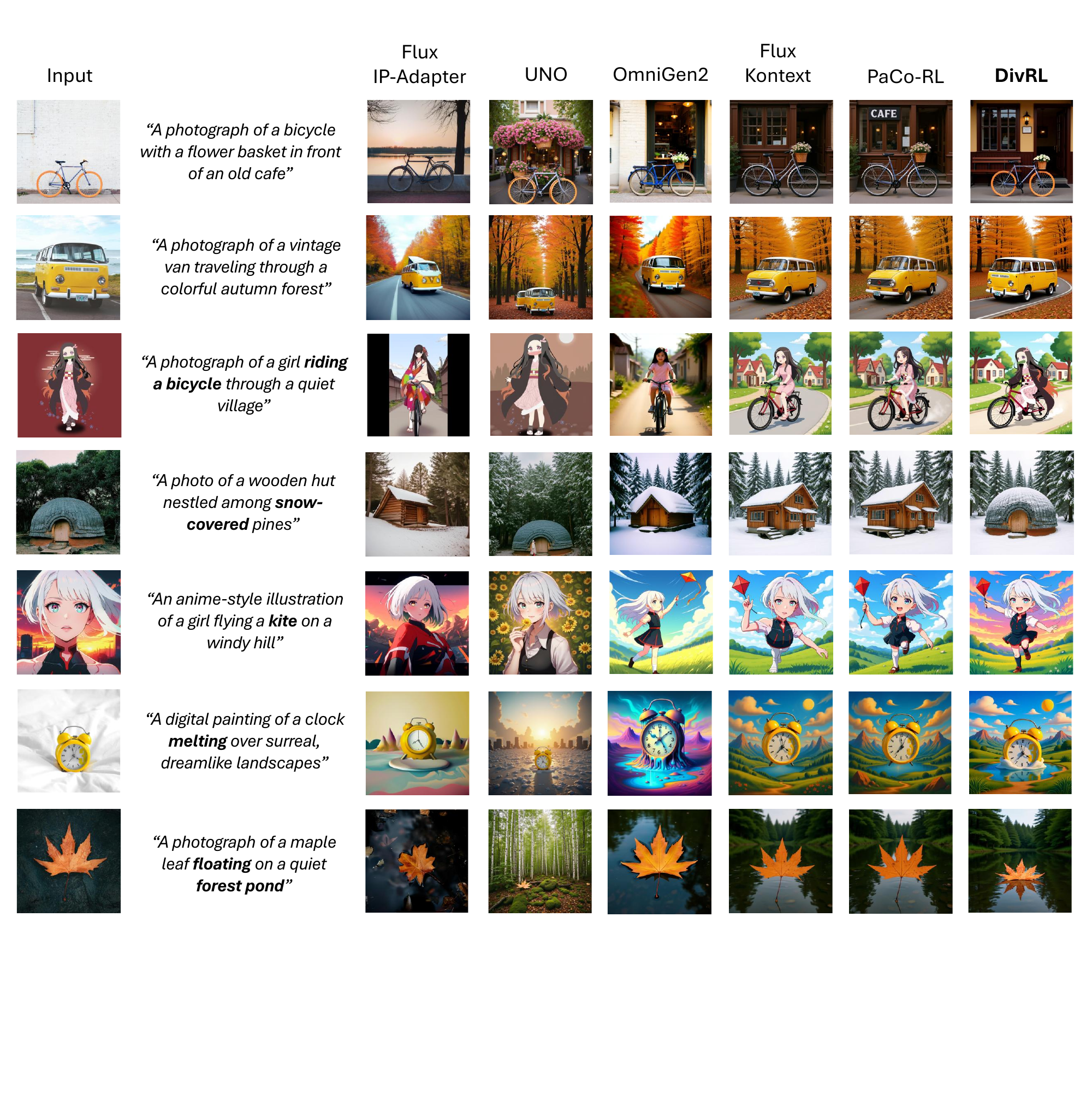}
\caption{Visual comparison with baselines. Our method produces structurally diverse images while preserving identity consistency and maintaining good prompt alignment.}
\label{fig:visual_comp}
\vspace{-4mm}
\end{figure}

\begin{figure}[t!]
\centering
\includegraphics[trim=0cm 3.5cm 0cm 0cm, width=0.8\textwidth]{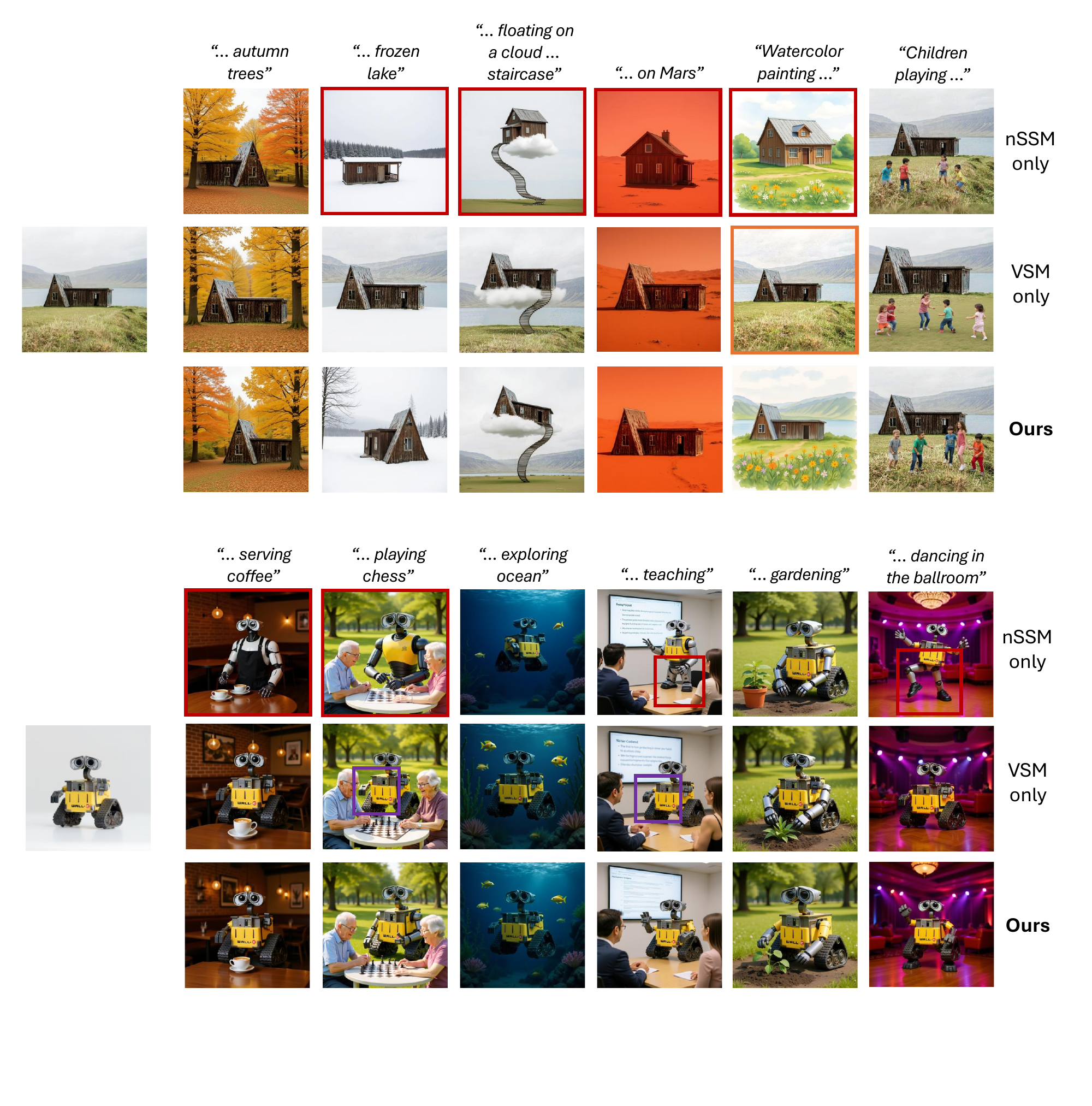}
\caption{Qualitative comparison between optimizing using nSSM only, VSM only, and ours. We highlight cases of \textcolor{red}{identity inconsistency}, \textcolor{orange}{prompt misalignment}, and \textcolor{purple}{structural rigidity}. Our method DivRL achieves the best balance between prompt following, identity preservation, and structural diversity. }
\label{fig:variants_comparison}
\vspace{-5mm}
\end{figure}

\begin{figure}[t]
\centering
    \includegraphics[width=0.9\linewidth]{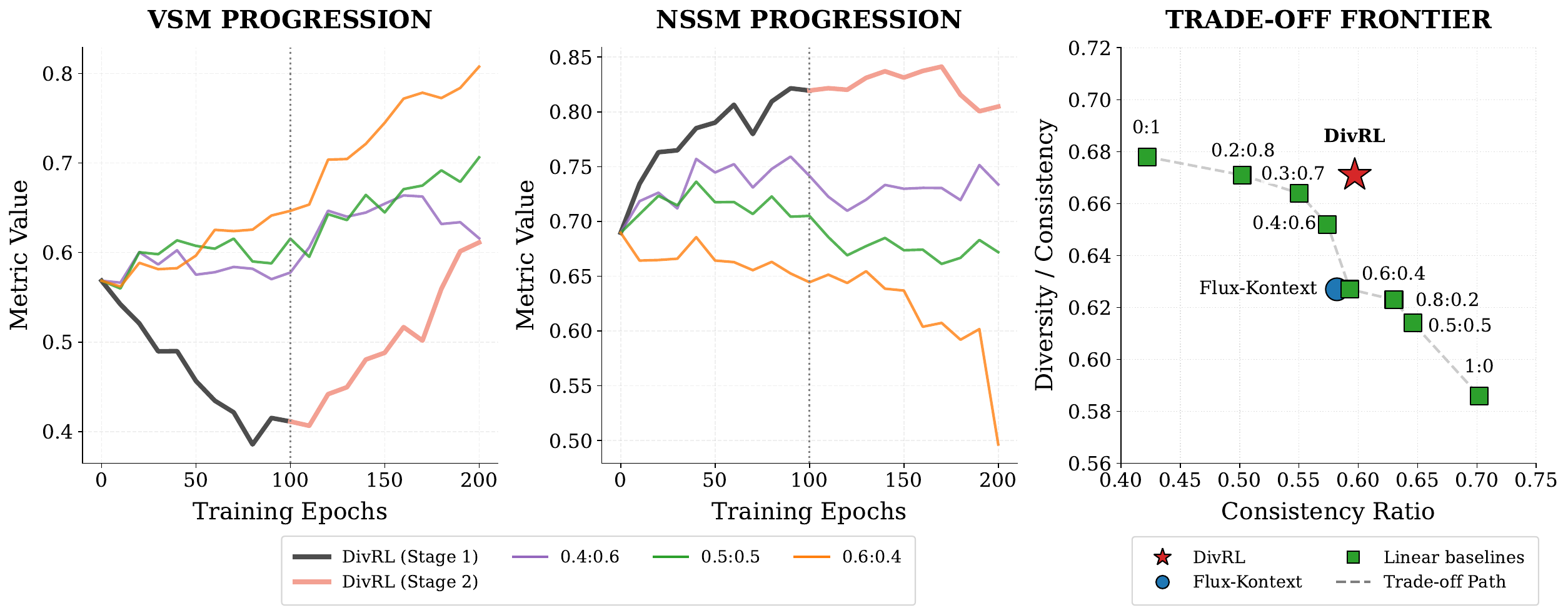}
    \caption{Comparison between our optimization strategy and linear reward weighting. Left is the training progression curve, while right is the trade-off frontier between the \textit{consistency ratio} and \textit{diversity-over-consistency}. For the linear weighting annotation, it is denoted as the linear ratio between VSM and nSSM.}
    \label{fig:linear_weight}
\end{figure}

\begin{figure}[h]
\vspace{-3mm}
\centering
\begin{minipage}[t]{0.2\columnwidth}
    \centering
    \includegraphics[trim=5cm 2.5cm 4.6cm 5cm, width=0.9\linewidth]{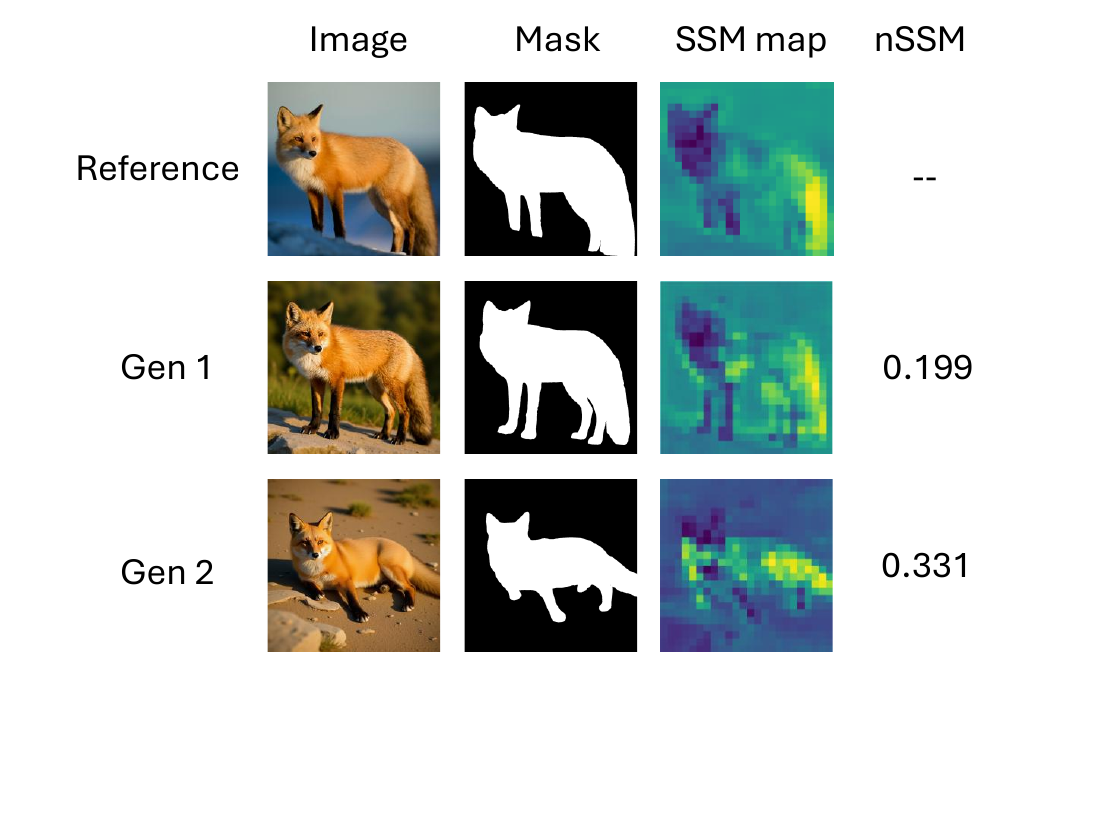}
    \caption{Visualization of the SSM maps and the corresponding nSSM scores.}
    \label{fig:nssm_vis}
\end{minipage}
\hfill
\begin{minipage}[t]{0.78\columnwidth}
    \centering
    \includegraphics[trim=0cm 5.5cm 0cm 0cm, width=0.9\textwidth]{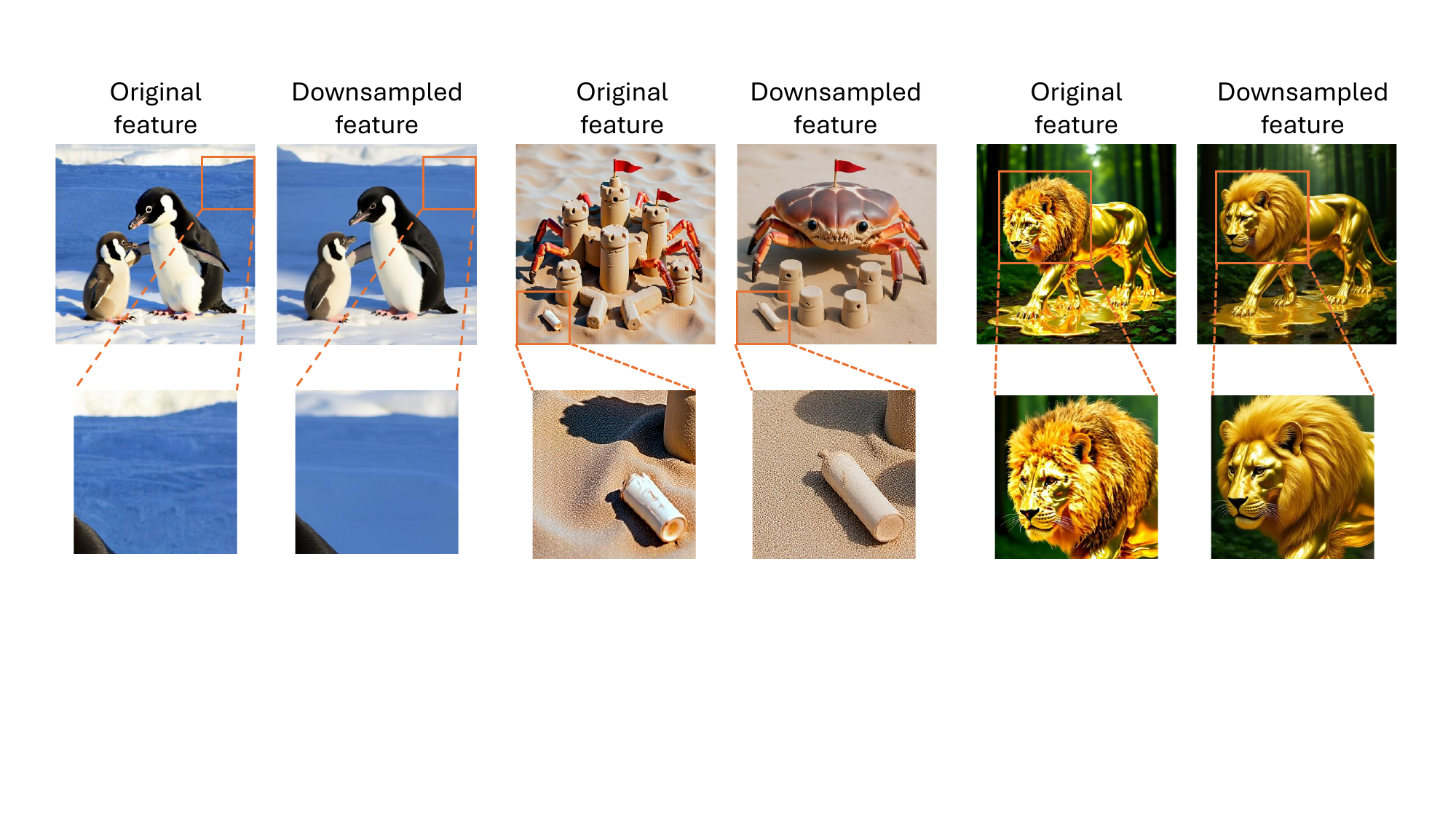}
    \caption{Comparison between computing nSSM using the original visual feature resolution and using downsampled features. Optimization with the original resolution introduces high-frequency artifacts, whereas downsampled features suppress these artifacts and produce visually pleasing results.}
    \label{fig:ssm_resolution}
\end{minipage}
\vspace{-3mm}
\end{figure}

\vspace{-4mm}
\subsection{Qualitative results}
We present qualitative comparisons between our method and the baselines in Figure~\ref{fig:visual_comp}. The examples include both rigid objects (\eg, bicycle and van) and articulated objects (\eg, stork and human). Our method preserves the fine visual details of the reference subject while following the text prompt and exhibiting structural diversity, such as changes in pose and viewpoint.

Flux-Kontext and PaCo-RL produce visually plausible results but often exhibit two failure modes. First, they occasionally fail to preserve fine-grained visual details. Second, although they maintain identity consistency, they sometimes fail to follow the prompt interactions. This behavior stems from the strong emphasis on identity preservation, which encourages the model to reproduce a subject configuration similar to the reference rather than adapting it to the contextual instructions. 

Other non–Flux-Kontext-based models tend to produce more structurally diverse outputs, which is consistent with the quantitative results. However, they exhibit different limitations. Flux-IP-Adapter often lacks fine visual details and overall fidelity. UNO generates visually consistent subjects but struggles with prompt alignment. OmniGen produces diverse and visually appealing images but frequently fails to preserve identity. These observations highlight the difficulty of simultaneously achieving strong identity preservation, structural diversity, prompt following, and high visual quality. Building upon the strong identity preservation capability of Flux-Kontext, our method relaxes the structural rigidity of the base model, enabling more diverse subject configurations while maintaining identity consistency and improving prompt alignment.

\vspace{-4mm}
\subsection{Ablation Studies}
\vspace{-2mm}
\subsubsection{Single-Reward Optimization}
We further compare our method with the two reward variants in Figure~\ref{fig:variants_comparison}. Optimizing with nSSM alone produces higher structural diversity by generating subjects with different poses and viewpoints, but it may alter the subject’s core attributes and lead to identity drift. In contrast, optimizing with VSM alone achieves strong identity preservation but may result in overly constrained outputs with poses and viewpoints similar to the reference image. In some cases, the model even preserves the original visual style of the reference, ignoring stylistic instructions from the prompt. By combining nSSM and VSM, our method achieves a better balance between identity consistency, structural diversity, and prompt following.

\vspace{-4mm}
\subsubsection{Optimization Strategy}
Our goal is to balance identity consistency and diversity during optimization. Since Flux-Kontext already provides strong identity preservation, we focus on improving diversity without degrading consistency. To analyze this trade-off, we compare our method with linear combinations of VSM and nSSM, as shown in Figure~\ref{fig:linear_weight}. As the training progression curves show, our method produces a modest but sufficient increase in VSM while achieving a substantial gain in nSSM over the Flux-Kontext baseline. Linear weighting, by contrast, usually fails to simultaneously improve both objectives. This can also be verified in the trade-off frontier graph that none of the linear weighting combinations can improve the \textit{consistency ratio} as well as the \textit{diversity-over-consistency} at the same time. This failure occurs because the gradients from the two reward signals are in direct competition under a linear combination. The gated formulation in our method decouples these two objectives: the model first explores structurally diverse solutions, and the identity gate selectively suppresses only those that drift beyond the consistency threshold, allowing VSM and nSSM to improve jointly.
\vspace{-5mm}
\subsubsection{nSSM visualization} 
We visualize the Self-Similarity Measure (SSM) maps in Fig~\ref{fig:nssm_vis}. The similarity scores within the masked region are aggregated to produce the SSM maps. 
When a generated image closely matches the reference structure (Gen 1), the SSM maps exhibit high similarity, resulting in a lower nSSM score. In contrast, a sample with a different pose (Gen 2) produces less similar SSM maps and therefore yields a higher nSSM score.
During training, masks are not generated for synthesized images; instead, nSSM is computed using features from all spatial locations of the generated images.
\vspace{-4mm}
\subsubsection{Spatial Resolution of the Self-Similarity Measure}

The MTG visual features have a spatial resolution of $48\times48$. While this resolution captures fine structural details, we found that directly computing nSSM at this scale leads to high-frequency artifacts during RL optimization. We hypothesize that optimizing structural similarity at such fine granularity encourages the model to reproduce dense local correlation patterns, allowing it to increase nSSM through high-frequency textures rather than meaningful structural changes, which is a form of reward hacking. To mitigate this issue, we apply $2\times2$ average pooling to downsample the feature maps to $24\times24$ before computing nSSM. As shown in Figure~\ref{fig:ssm_resolution}, the downsampled features effectively suppress high-frequency artifacts and produce visually cleaner results.

\vspace{-4mm}
\section{Discussion and Limitations}
\vspace{-2mm}
Our results highlight an important observation in subject-driven generation: identity preservation rewards alone implicitly encourage structural mimicry. Our framework addresses this issue by explicitly separating structural diversity from identity preservation. An interesting side effect of encouraging structural diversity is the improvement in prompt interaction. As shown in our experiments, relaxing structural rigidity allows the model to better adapt the subject to contextual cues in the prompt.  

\noindent \textbf{Limitations.} Despite these advantages, several limitations were observed.
Firstly, the proposed diversity metric focuses on structural variations in pose and viewpoint, and may not fully capture higher-level semantic diversity or compositional changes. 
Secondly, our approach relies on RL for post-training, which introduces additional computational overhead compared to purely feed-forward methods. 
Finally, our base model Flux-Kontext accepts a single reference image by default and may be less effective when the reference contains ambiguous identity cues or complex occlusions. 
Addressing these limitations through richer diversity metrics, more efficient optimization strategies, and multi-reference conditioning is an interesting direction for future work.
\vspace{-2mm}
\section{Conclusions}
\vspace{-2mm}
In this paper, we investigate the identity–diversity trade-off in subject-driven image generation. We identify structural mimicry as a common failure mode in existing methods, where models tend to reproduce the spatial configuration of the reference image instead of generating diverse structures. To address this challenge, we propose a RL-based framework that decouples structural diversity from identity preservation. We introduce a negative Self-Similarity Measure (nSSM) to quantify structural diversity and employ Visual Semantic Matching (VSM) as an identity consistency gate. Combined with an Explore-and-Suppress optimization strategy based on Flow-GRPO, this formulation encourages the model to explore diverse structural configurations while filtering identity-inconsistent samples. Experiments demonstrate that the proposed method improves structural diversity while maintaining strong identity consistency, achieving a better balance between diversity, prompt following, and identity preservation compared with existing approaches.

\bibliographystyle{splncs04}
\bibliography{main}

@String(CVPR  = {IEEE Conf. Comput. Vis. Pattern Recog.})

@String(ICCV  = {Int. Conf. Comput. Vis.})

@String(CVPR  = {CVPR})

@String(ICCV  = {ICCV})

@misc{deepseek-math,
      title={DeepSeekMath: Pushing the Limits of Mathematical Reasoning in Open Language Models}, 
      author={Zhihong Shao and Peiyi Wang and Qihao Zhu and Runxin Xu and Junxiao Song and Xiao Bi and Haowei Zhang and Mingchuan Zhang and Y. K. Li and Y. Wu and Daya Guo},
      year={2024},
      eprint={2402.03300},
      archivePrefix={arXiv},
      primaryClass={cs.CL},
      url={https://arxiv.org/abs/2402.03300}, 
}

@inproceedings{
liu2025flowgrpo,
title={Flow-{GRPO}: Training Flow Matching Models via Online {RL}},
author={Jie Liu and Gongye Liu and Jiajun Liang and Yangguang Li and Jiaheng Liu and Xintao Wang and Pengfei Wan and Di ZHANG and Wanli Ouyang},
booktitle={The Thirty-ninth Annual Conference on Neural Information Processing Systems},
year={2025},
url={https://openreview.net/forum?id=oCBKGw5HNf}
}

@inproceedings{eldesokey2025mindtheglitch,
  title={Mind-the-Glitch: Visual Correspondence for Detecting Inconsistencies in Subject-Driven Generation},
  author={Eldesokey, Abdelrahman and Cvejic, Aleksandar and Ghanem, Bernard and Wonka, Peter},
  booktitle={Advances in Neural Information Processing Systems},
  year={2025}
}

@inproceedings{
gal2023textualinversion,
title={An Image is Worth One Word: Personalizing Text-to-Image Generation using Textual Inversion},
author={Rinon Gal and Yuval Alaluf and Yuval Atzmon and Or Patashnik and Amit Haim Bermano and Gal Chechik and Daniel Cohen-Or},
booktitle={The Eleventh International Conference on Learning Representations },
year={2023},
url={https://openreview.net/forum?id=NAQvF08TcyG}
}

@inproceedings{ruiz2023dreambooth,
  title={Dreambooth: Fine tuning text-to-image diffusion models for subject-driven generation},
  author={Ruiz, Nataniel and Li, Yuanzhen and Jampani, Varun and Pritch, Yael and Rubinstein, Michael and Aberman, Kfir},
  booktitle={Proceedings of the IEEE/CVF Conference on Computer Vision and Pattern Recognition},
  year={2023}
}

@article{ye2023ip-adapter,
  title={IP-Adapter: Text Compatible Image Prompt Adapter for Text-to-Image Diffusion Models},
  author={Ye, Hu and Zhang, Jun and Liu, Sibo and Han, Xiao and Yang, Wei},
  booktitle={arXiv preprint arxiv:2308.06721},
  year={2023}
}

@inproceedings{
li2023blipdiffusion,
title={{BLIP}-Diffusion: Pre-trained Subject Representation for Controllable Text-to-Image Generation and Editing},
author={Dongxu Li and Junnan Li and Steven Hoi},
booktitle={Thirty-seventh Conference on Neural Information Processing Systems},
year={2023},
url={https://openreview.net/forum?id=g6We1SwaY9}
}

@misc{zhang2023controlnet,
  title={Adding Conditional Control to Text-to-Image Diffusion Models}, 
  author={Lvmin Zhang and Anyi Rao and Maneesh Agrawala},
  booktitle={IEEE International Conference on Computer Vision (ICCV)},
  year={2023},
}

@misc{labs2025flux1kontext,
      title={FLUX.1 Kontext: Flow Matching for In-Context Image Generation and Editing in Latent Space}, 
      author={Black Forest Labs and Stephen Batifol and Andreas Blattmann and Frederic Boesel and Saksham Consul and Cyril Diagne and Tim Dockhorn and Jack English and Zion English and Patrick Esser and Sumith Kulal and Kyle Lacey and Yam Levi and Cheng Li and Dominik Lorenz and Jonas Müller and Dustin Podell and Robin Rombach and Harry Saini and Axel Sauer and Luke Smith},
      year={2025},
      eprint={2506.15742},
      archivePrefix={arXiv},
      primaryClass={cs.GR},
      url={https://arxiv.org/abs/2506.15742}, 
}

@InProceedings{Xiao2025omnigen,
    author    = {Xiao, Shitao and Wang, Yueze and Zhou, Junjie and Yuan, Huaying and Xing, Xingrun and Yan, Ruiran and Li, Chaofan and Wang, Shuting and Huang, Tiejun and Liu, Zheng},
    title     = {OmniGen: Unified Image Generation},
    booktitle = {Proceedings of the IEEE/CVF Conference on Computer Vision and Pattern Recognition (CVPR)},
    month     = {June},
    year      = {2025},
    pages     = {13294-13304}
}

@article{wu2025omnigen2,
  title={OmniGen2: Exploration to Advanced Multimodal Generation},
  author={Chenyuan Wu and Pengfei Zheng and Ruiran Yan and Shitao Xiao and Xin Luo and Yueze Wang and Wanli Li and Xiyan Jiang and Yexin Liu and Junjie Zhou and Ze Liu and Ziyi Xia and Chaofan Li and Haoge Deng and Jiahao Wang and Kun Luo and Bo Zhang and Defu Lian and Xinlong Wang and Zhongyuan Wang and Tiejun Huang and Zheng Liu},
  journal={arXiv preprint arXiv:2506.18871},
  year={2025}
}

@misc{wu2025qwenimage,
      title={Qwen-Image Technical Report}, 
      author={Chenfei Wu and Jiahao Li and Jingren Zhou and Junyang Lin and Kaiyuan Gao and Kun Yan and Sheng-ming Yin and Shuai Bai and Xiao Xu and Yilei Chen and Yuxiang Chen and Zecheng Tang and Zekai Zhang and Zhengyi Wang and An Yang and Bowen Yu and Chen Cheng and Dayiheng Liu and Deqing Li and Hang Zhang and Hao Meng and Hu Wei and Jingyuan Ni and Kai Chen and Kuan Cao and Liang Peng and Lin Qu and Minggang Wu and Peng Wang and Shuting Yu and Tingkun Wen and Wensen Feng and Xiaoxiao Xu and Yi Wang and Yichang Zhang and Yongqiang Zhu and Yujia Wu and Yuxuan Cai and Zenan Liu},
      year={2025},
      eprint={2508.02324},
      archivePrefix={arXiv},
      primaryClass={cs.CV},
      url={https://arxiv.org/abs/2508.02324}, 
}

@inproceedings{
he2026tempflowgrpo,
title={{TEMPFLOW}-{GRPO}: {WHEN} {TIMING} {MATTERS} {FOR} {GRPO} {IN} {FLOW} {MODELS}},
author={Xiaoxuan He and Siming Fu and Yuke Zhao and Wanli Li and Jian Yang and Dacheng Yin and Fengyun Rao and Bo Zhang},
booktitle={The Fourteenth International Conference on Learning Representations},
year={2026},
url={https://openreview.net/forum?id=7mCo3R3Wyn}
}

@inproceedings{xu2023imagereward,
  title={ImageReward: learning and evaluating human preferences for text-to-image generation},
  author={Xu, Jiazheng and Liu, Xiao and Wu, Yuchen and Tong, Yuxuan and Li, Qinkai and Ding, Ming and Tang, Jie and Dong, Yuxiao},
  booktitle={Proceedings of the 37th International Conference on Neural Information Processing Systems},
  pages={15903--15935},
  year={2023}
}

@InProceedings{Wallace2024diffusiondpo,
    author    = {Wallace, Bram and Dang, Meihua and Rafailov, Rafael and Zhou, Linqi and Lou, Aaron and Purushwalkam, Senthil and Ermon, Stefano and Xiong, Caiming and Joty, Shafiq and Naik, Nikhil},
    title     = {Diffusion Model Alignment Using Direct Preference Optimization},
    booktitle = {Proceedings of the IEEE/CVF Conference on Computer Vision and Pattern Recognition (CVPR)},
    month     = {June},
    year      = {2024},
    pages     = {8228-8238}
}

@inproceedings{rombach2022sd,
  title={High-resolution image synthesis with latent diffusion models},
  author={Rombach, Robin and Blattmann, Andreas and Lorenz, Dominik and Esser, Patrick and Ommer, Bj{\"o}rn},
  booktitle={Proceedings of the IEEE/CVF conference on computer vision and pattern recognition},
  pages={10684--10695},
  year={2022}
}

@inproceedings{esser2024sd3,
  title={Scaling rectified flow transformers for high-resolution image synthesis},
  author={Esser, Patrick and Kulal, Sumith and Blattmann, Andreas and Entezari, Rahim and M{\"u}ller, Jonas and Saini, Harry and Levi, Yam and Lorenz, Dominik and Sauer, Axel and Boesel, Frederic and others},
  booktitle={Forty-first international conference on machine learning},
  year={2024}
}

@inproceedings{sauer2024fast,
  title={Fast high-resolution image synthesis with latent adversarial diffusion distillation},
  author={Sauer, Axel and Boesel, Frederic and Dockhorn, Tim and Blattmann, Andreas and Esser, Patrick and Rombach, Robin},
  booktitle={SIGGRAPH Asia 2024 Conference Papers},
  pages={1--11},
  year={2024}
}

@inproceedings{chen2024anydoor,
  title={Anydoor: Zero-shot object-level image customization},
  author={Chen, Xi and Huang, Lianghua and Liu, Yu and Shen, Yujun and Zhao, Deli and Zhao, Hengshuang},
  booktitle={Proceedings of the IEEE/CVF conference on computer vision and pattern recognition},
  pages={6593--6602},
  year={2024}
}

@article{liu2025tuna,
  title={Tuna: Taming unified visual representations for native unified multimodal models},
  author={Liu, Zhiheng and Ren, Weiming and Liu, Haozhe and Zhou, Zijian and Chen, Shoufa and Qiu, Haonan and Huang, Xiaoke and An, Zhaochong and Yang, Fanny and Patel, Aditya and others},
  journal={arXiv preprint arXiv:2512.02014},
  year={2025}
}

@article{chen2025blip3,
  title={Blip3-o: A family of fully open unified multimodal models-architecture, training and dataset},
  author={Chen, Jiuhai and Xu, Zhiyang and Pan, Xichen and Hu, Yushi and Qin, Can and Goldstein, Tom and Huang, Lifu and Zhou, Tianyi and Xie, Saining and Savarese, Silvio and others},
  journal={arXiv preprint arXiv:2505.09568},
  year={2025}
}

@article{kirstain2023pickscore,
  title={Pick-a-pic: An open dataset of user preferences for text-to-image generation},
  author={Kirstain, Yuval and Polyak, Adam and Singer, Uriel and Matiana, Shahbuland and Penna, Joe and Levy, Omer},
  journal={Advances in neural information processing systems},
  volume={36},
  pages={36652--36663},
  year={2023}
}

@article{wu2023hps,
  title={Human preference score v2: A solid benchmark for evaluating human preferences of text-to-image synthesis},
  author={Wu, Xiaoshi and Hao, Yiming and Sun, Keqiang and Chen, Yixiong and Zhu, Feng and Zhao, Rui and Li, Hongsheng},
  journal={arXiv preprint arXiv:2306.09341},
  year={2023}
}

@inproceedings{lin2024vqascore,
  title={Evaluating text-to-visual generation with image-to-text generation},
  author={Lin, Zhiqiu and Pathak, Deepak and Li, Baiqi and Li, Jiayao and Xia, Xide and Neubig, Graham and Zhang, Pengchuan and Ramanan, Deva},
  booktitle={European Conference on Computer Vision},
  pages={366--384},
  year={2024},
  organization={Springer}
}

@inproceedings{
zhu2025dspo,
title={{DSPO}: Direct Score Preference Optimization for Diffusion Model Alignment},
author={Huaisheng Zhu and Teng Xiao and Vasant G Honavar},
booktitle={The Thirteenth International Conference on Learning Representations},
year={2025},
url={https://openreview.net/forum?id=xyfb9HHvMe}
}

@inproceedings{na2025boost,
  title={Boost your human image generation model via direct preference optimization},
  author={Na, Sanghyeon and Kim, Yonggyu and Lee, Hyunjoon},
  booktitle={Proceedings of the Computer Vision and Pattern Recognition Conference},
  pages={23551--23562},
  year={2025}
}

@inproceedings{shi2024instantbooth,
  title={Instantbooth: Personalized text-to-image generation without test-time finetuning},
  author={Shi, Jing and Xiong, Wei and Lin, Zhe and Jung, Hyun Joon},
  booktitle={Proceedings of the IEEE/CVF conference on computer vision and pattern recognition},
  pages={8543--8552},
  year={2024}
}

@inproceedings{black2023ddpo,
      title={Training Diffusion Models with Reinforcement Learning},
      author={Kevin Black and Michael Janner and Yilun Du and Ilya Kostrikov and Sergey Levine},
      year={2023},
      eprint={2305.13301},
      archivePrefix={arXiv},
      primaryClass={cs.LG}
}

@inproceedings{miao2024training,
  title={Training diffusion models towards diverse image generation with reinforcement learning},
  author={Miao, Zichen and Wang, Jiang and Wang, Ze and Yang, Zhengyuan and Wang, Lijuan and Qiu, Qiang and Liu, Zicheng},
  booktitle={Proceedings of the IEEE/CVF Conference on Computer Vision and Pattern Recognition},
  pages={10844--10853},
  year={2024}
}

@article{ouyang2022rlhf,
  title={Training language models to follow instructions with human feedback},
  author={Ouyang, Long and Wu, Jeffrey and Jiang, Xu and Almeida, Diogo and Wainwright, Carroll and Mishkin, Pamela and Zhang, Chong and Agarwal, Sandhini and Slama, Katarina and Ray, Alex and others},
  journal={Advances in neural information processing systems},
  volume={35},
  pages={27730--27744},
  year={2022}
}

@article{fan2023dpok,
  title={Dpok: Reinforcement learning for fine-tuning text-to-image diffusion models},
  author={Fan, Ying and Watkins, Olivia and Du, Yuqing and Liu, Hao and Ryu, Moonkyung and Boutilier, Craig and Abbeel, Pieter and Ghavamzadeh, Mohammad and Lee, Kangwook and Lee, Kimin},
  journal={Advances in Neural Information Processing Systems},
  volume={36},
  pages={79858--79885},
  year={2023}
}

@article{zheng2023secrets,
  title={Secrets of rlhf in large language models part i: Ppo},
  author={Zheng, Rui and Dou, Shihan and Gao, Songyang and Hua, Yuan and Shen, Wei and Wang, Binghai and Liu, Yan and Jin, Senjie and Liu, Qin and Zhou, Yuhao and others},
  journal={arXiv preprint arXiv:2307.04964},
  year={2023}
}

@article{xue2025dancegrpo,
  title={Dancegrpo: Unleashing grpo on visual generation},
  author={Xue, Zeyue and Wu, Jie and Gao, Yu and Kong, Fangyuan and Zhu, Lingting and Chen, Mengzhao and Liu, Zhiheng and Liu, Wei and Guo, Qiushan and Huang, Weilin and others},
  journal={arXiv preprint arXiv:2505.07818},
  year={2025}
}

@article{li2025mixgrpo,
  title={Mixgrpo: Unlocking flow-based grpo efficiency with mixed ode-sde},
  author={Li, Junzhe and Cui, Yutao and Huang, Tao and Ma, Yinping and Fan, Chun and Yang, Miles and Zhong, Zhao},
  journal={arXiv preprint arXiv:2507.21802},
  year={2025}
}

@article{luo2025editscore,
  title={Editscore: Unlocking online rl for image editing via high-fidelity reward modeling},
  author={Luo, Xin and Wang, Jiahao and Wu, Chenyuan and Xiao, Shitao and Jiang, Xiyan and Lian, Defu and Zhang, Jiajun and Liu, Dong and others},
  journal={arXiv preprint arXiv:2509.23909},
  year={2025}
}

@article{wu2025editreward,
  title={Editreward: A human-aligned reward model for instruction-guided image editing},
  author={Wu, Keming and Jiang, Sicong and Ku, Max and Nie, Ping and Liu, Minghao and Chen, Wenhu},
  journal={arXiv preprint arXiv:2509.26346},
  year={2025}
}

@article{long2026spatialreward,
  title={SpatialReward: Bridging the Perception Gap in Online RL for Image Editing via Explicit Spatial Reasoning},
  author={Long, Yancheng and Yang, Yankai and Wei, Hongyang and Chen, Wei and Zhang, Tianke and Liu, Changyi and Jiang, Kaiyu and Chen, Jiankang and Tang, Kaiyu and Wen, Bin and others},
  journal={arXiv preprint arXiv:2602.07458},
  year={2026}
}

@article{ping2025paco,
  title={PaCo-RL: Advancing Reinforcement Learning for Consistent Image Generation with Pairwise Reward Modeling},
  author={Ping, Bowen and Jia, Chengyou and Luo, Minnan and Xia, Changliang and Shen, Xin and Dang, Zhuohang and Qian, Hangwei},
  journal={arXiv preprint arXiv:2512.04784},
  year={2025}
}

@article{meng2025identitygrpo,
  title={Identity-grpo: Optimizing multi-human identity-preserving video generation via reinforcement learning},
  author={Meng, Xiangyu and Zhang, Zixian and Zhang, Zhenghao and Liao, Junchao and Qin, Long and Wang, Weizhi},
  journal={arXiv preprint arXiv:2510.14256},
  year={2025}
}

@inproceedings{hu2025towards,
  title={Towards better alignment: Training diffusion models with reinforcement learning against sparse rewards},
  author={Hu, Zijing and Zhang, Fengda and Chen, Long and Kuang, Kun and Li, Jiahui and Gao, Kaifeng and Xiao, Jun and Wang, Xin and Zhu, Wenwu},
  booktitle={Proceedings of the Computer Vision and Pattern Recognition Conference},
  pages={23604--23614},
  year={2025}
}

@inproceedings{
goyal2025shortcut,
title={Preventing Shortcuts in Adapter Training via Providing the Shortcuts},
author={Anujraaj Goyal and Guocheng Qian and Huseyin Coskun and Aarush Gupta and Himmy Tam and Daniil Ostashev and Ju Hu and Dhritiman Sagar and Sergey Tulyakov and Kfir Aberman and Kuan-Chieh Wang},
booktitle={The Thirty-ninth Annual Conference on Neural Information Processing Systems},
year={2025},
url={https://openreview.net/forum?id=cZMno8E3yp}
}

@inproceedings{kumari2025syncd,
  title={Generating Multi-Image Synthetic Data for Text-to-Image Customization},
  author={Kumari, Nupur and Yin, Xi and Zhu, Jun-Yan and Misra, Ishan and Azadi, Samaneh},
  booktitle={IEEE International Conference on Computer Vision (ICCV)},
  year={2025}
}

@inproceedings{wu2025uno,
  title={Less-to-more generalization: Unlocking more controllability by in-context generation},
  author={Wu, Shaojin and Huang, Mengqi and Wu, Wenxu and Cheng, Yufeng and Ding, Fei and He, Qian},
  booktitle={Proceedings of the IEEE/CVF International Conference on Computer Vision},
  pages={18682--18692},
  year={2025}
}

@inproceedings{peng2024dreambench_plus,
    author={Yuang Peng and Yuxin Cui and Haomiao Tang and Zekun Qi and Runpei Dong and Jing Bai and Chunrui Han and Zheng Ge and Xiangyu Zhang and Shu-Tao Xia},
    title={DreamBench++: A Human-Aligned Benchmark for Personalized Image Generation},
    booktitle={The Thirteenth International Conference on Learning Representations},
    year={2025},
    url={https://dreambenchplus.github.io/},
}

\appendix

\section{Evaluation metrics}
\subsection{Structural DINO (sDINO)} sDINO measures the structural similarity between the patches across the reference image $I_{ref}$ and the generated image $I_{gen}$. It focuses on the local feature correspondence. Specifically, let $G = {g_1, g_2, ..., g_n}$ be the set of $n$ patch embeddings for the generated image, and $R = {r_1, r_2, ..., r_m}$ be the set of $m$ patch embeddings for the reference image. The Structural DINO (sDINO) score is defined as the mean of the maximum cosine similarities between each generated patch and the entire set of reference patches:
\begin{align}
\text{sDINO}(G, R) = \frac{1}{n}\sum\limits_{i=1}^{n} \max\limits_{j\in \{1,..., m \}}\left( \frac{g_i \cdot r_j}{\|g_i\| \|r_j\|} \right)
\end{align}

\subsection{Scale-Invariant IoU} Scale-Invariant Intersection over Union (si-IoU) measures the shape similarity between the segmentation masks of the generated subject and the reference subject by decoupling their relative sizes and positions. As we already have the ground-truth reference image mask $M_{r}$, we first use an off-the-shelf segmentation model to extract the segmentation mask $M_s$ for the generated subject. Then, we crop the regions of both masks to get the tightest axis-aligned bounding boxes containing all non-zero pixels $\widetilde{M}_r$ and $\widetilde{M}_g$, respectively. We follow by reshaping $\widetilde{M}_g$ to $\widehat{M}_g$ in order to match the spatial resolution of $\widetilde{M}_r$. Finally, we compute the IoU between $\widehat{M}_g$ and $\widetilde{M}_r$. Intuitively, a higher number of si-IoU indicates a higher structural similarity between the generated subject and the reference subject. 

\section{Implementation Details: Computation, Metrics and Evaluation}
We train our model for 48 hours on 8 A100 80GB GPUs, with each optimization stage taking 24 hours. The LoRA weights add 1.8\% parameters to the base Flux-Kontext model.  
Our inference time to generate one image under 28 denoising steps is 34s on 1 A100 40GB GPU, which is the same cost as the base model. 

To evaluate subject-driven generation performance, we employ a suite of vision-language and structural similarity metrics. For CLIP Text (CLIP-T) and CLIP Image (CLIP-I) similarity, we use ViT-L/14 as the backbone. In line with standard practice, CLIP-I is calculated using the full reference and generated images without object masking. Similarly, DINO cosine similarity is computed using the full images, but with DINOv2-base as the base model.

For finer structural evaluation, specifically DINO-nSSM, MTG-nSSM, and sDINO, we utilize DINOv2-small as the backbone. For these metrics, we extract features exclusively from the subject regions using subject masks for both reference and generated images. Masks for the generated images are generated during the evaluation stage using the off-the-shelf Grounded-Segment-Anything~\footnote{\href{https://github.com/IDEA-Research/Grounded-Segment-Anything}{https://github.com/IDEA-Research/Grounded-Segment-Anything}} model. Notably, our observations indicate that results remain consistent whether or not features outside the generated subject mask are excluded.

When calculating VSM and MTG-nSSM, we omit the ``style'' sub-category from the DreamBench++ benchmark because style transfer is fundamentally distinct from subject-driven generation. Since the subject of interest often exists only in the text prompt and not the reference style image, establishing meaningful correspondences for VSM calculation; however, to maintain consistency with prior work, the ``style'' sub-category is included for all other evaluation metrics. 

Regarding the resolution of the visual features map, we use average pooling to downsample the visual feature maps from $48 \times 48$ to $24 \times 24$ during training to mitigate high-frequency artifacts, as discussed in our ablation study in the main paper. During evaluation, we maintain the original $48 \times 48$ resolution to ensure the metrics remain sensitive to fine-grained structural changes and intricate details.

\section{Backbone Generalization}

The nSSM reward is computed over any dense visual feature grid, and the two-stage gated optimization is compatible with any differentiable identity consistency metric. This makes the framework backbone-agnostic: when using a different feature extractor, nSSM is computed using that backbone's patch features, while VSM is replaced by an appropriate similarity metric for the chosen backbone. To verify this, we replace MTG features with DINOv2 features and substitute the cosine similarity as the identity gate, re-running the full two-stage optimization with the same hyperparameters. As shown in Table~\ref{tab:backbone_ablation}, the DINO-backbone variant achieves performance comparable to the MTG-backbone default across all metrics, with a slight improvement in consistency ratio and diversity-over-consistency. This confirms that the Explore-and-Suppress framework can generalize beyond the default MTG feature backbone.

\begin{table}[h]
\centering
\small
\caption{Backbone generalization: our method using MTG vs.\ DINOv2 visual features.}
\label{tab:backbone_ablation}
\renewcommand{\arraystretch}{1.2}
\setlength{\tabcolsep}{4pt}
\resizebox{\textwidth}{!}{
\begin{tabular}{l c | ccc | ccc | c}
\toprule
& \textbf{Prompt foll.} & \multicolumn{3}{c|}{\textbf{Identity consistency}} & \multicolumn{3}{c|}{\textbf{Diversity}} & \textbf{Aesthetic} \\
\textbf{Model} & CLIP-T$\uparrow$ & CLIP-I$\uparrow$ & DINO$\uparrow$ & VSM-0.7$\uparrow$ & DINO-nSSM$\uparrow$ & Scale-inv IOU$\downarrow$ & MTG-nSSM$\uparrow$ & HPS$\uparrow$ \\
\midrule
Flux-Kontext                   & 0.283 & 0.781 & 0.594 & 0.605 & 0.433 & 0.694 & 0.659 & 0.294 \\
\rowcolor{highlight} Ours (MTG backbone) & 0.279 & 0.786 & 0.600 & 0.614 & 0.453 & 0.697 & 0.689 & 0.287 \\
\rowcolor{highlight} Ours (DINO backbone) & 0.280 & 0.780 & 0.595 & 0.602 & 0.455 & 0.693 & 0.698 & 0.297 \\
\bottomrule
\end{tabular}}
\end{table}

\section{Ablation study}
We provide an ablation on the threshold $s$ for the hinge loss during Stage 2 of the optimization. We report the comprehensive quantitative results in Table~\ref{tab:s_ablation}. In addition, we also include OmniGen2~\cite{wu2025omnigen2} for comparison. We show that $s$ has a noticeable impact on the performance of the models. Optimizing with $s = 0.7$ will result in the best structural diversity. However, the identity consistency has a slight drop. Optimizing with $s = 0.4$ will result in the best diversity, with a cost of lower identity consistency than the baseline. 
We empirically find out that using $s = 0.5$ can achieve the best balance between the identity consistency and diversity, so we use it as our default setting. From the qualitative results, we observed that both $s=0.5$ and $0.6$ obtain the best results and both can be used in practice.

We provide another ablation on the weighting term $\lambda$ during the second stage of optimization in Table~\ref{tab:lambda_ablation}. Using $\lambda=0.5$ applies a weak penalty on identity drift; the model has a weak penalty of identity drift, resulting in high diversity but low VSM. Conversely, using $\lambda=50$ produces a counterintuitive result: despite the stronger penalty, VSM is lower than with $\lambda=5$. We attribute this to optimization instability. When the penalty coefficient is very large, any sample with $\text{VSM} < s$ generates an extremely large gradient signal. This causes destabilizing parameter updates that prevent stable convergence. $\lambda=5$ strikes the right balance: firm enough to suppress identity-inconsistent samples, but not so large as to destabilize optimization.


\begin{table}[t]
\centering
\small
\caption{Ablation on the threshold $s$.}
\label{tab:s_ablation}
\renewcommand{\arraystretch}{1.2} 
\setlength{\tabcolsep}{4pt}      
\resizebox{\textwidth}{!}{
\begin{tabular}{l c | ccc | ccc | c} 
\toprule
& \textbf{Prompt foll.} & \multicolumn{3}{c|}{\textbf{Identity consistency}} & \multicolumn{3}{c|}{\textbf{Diversity}} & \textbf{Aesthetic} \\ 
\textbf{Model} & CLIP-T$\uparrow$ & CLIP-I$\uparrow$ & DINO$\uparrow$ & VSM-0.7$\uparrow$ & DINO-nSSM$\uparrow$ & Scale-inv IOU$\downarrow$ & MTG-nSSM$\uparrow$ & HPS$\uparrow$ \\
\midrule
Flux-Kontext    & 0.283 & 0.781 & 0.594 & 0.605 & 0.433 & 0.694 & 0.659 & 0.294 \\
\rowcolor{highlight} $s = 0.4$ & 0.282 & 0.761 & 0.527 & 0.534 & \textbf{0.487} & \textbf{0.652} & \textbf{0.713} & 0.297 \\
\rowcolor{highlight} $s = 0.5$ & 0.279 & \textbf{0.786} & \textbf{0.600} & \textbf{0.614} & 0.453 & 0.697 & 0.689 & 0.287 \\
\rowcolor{highlight} $s = 0.6$ & 0.284 & 0.781 & 0.592 & 0.596 & 0.446 & 0.685 & 0.671 & 0.294 \\
\rowcolor{highlight} $s = 0.7$ & \textbf{0.286} & 0.773 & 0.560 & 0.570 & 0.476 & 0.663 & 0.707 & \textbf{0.301} \\
\bottomrule
\end{tabular}}
\end{table}

\begin{table}[t]
\centering
\small
\caption{Ablation on the weighting parameter $\lambda$.}
\label{tab:lambda_ablation}
\renewcommand{\arraystretch}{1.2} 
\setlength{\tabcolsep}{4pt}      
\resizebox{\textwidth}{!}{
\begin{tabular}{l c | ccc | ccc | c} 
\toprule
& \textbf{Prompt foll.} & \multicolumn{3}{c|}{\textbf{Identity consistency}} & \multicolumn{3}{c|}{\textbf{Diversity}} & \textbf{Aesthetic} \\ 
\textbf{Model} & CLIP-T$\uparrow$ & CLIP-I$\uparrow$ & DINO$\uparrow$ & VSM-0.7$\uparrow$ & DINO-nSSM$\uparrow$ & Scale-inv IOU$\downarrow$ & MTG-nSSM$\uparrow$ & HPS$\uparrow$ \\
\midrule
Flux-Kontext    & 0.283 & 0.781 & 0.594 & 0.605 & 0.433 & 0.694 & 0.659 & 0.294 \\
\rowcolor{highlight} $\lambda=0.5$ & \textbf{0.286} & 0.764 & 0.537 & 0.535 & \textbf{0.487} & \textbf{0.727} & \textbf{0.724} & 0.297 \\
\rowcolor{highlight} $\lambda=5$ & 0.279 & \textbf{0.786} & \textbf{0.600} & \textbf{0.614} & 0.453 & 0.697 & 0.689 & 0.287 \\
\rowcolor{highlight} $\lambda=50$ & 0.283 & 0.780 & 0.583 & 0.580 & 0.469 & 0.716 & 0.689 & 0.293 \\
\bottomrule
\end{tabular}}
\end{table}

\section{More qualitative results}
\subsection{Multi-prompt comparison}

\begin{figure}[t!]
\centering
\includegraphics[width=0.9\textwidth]{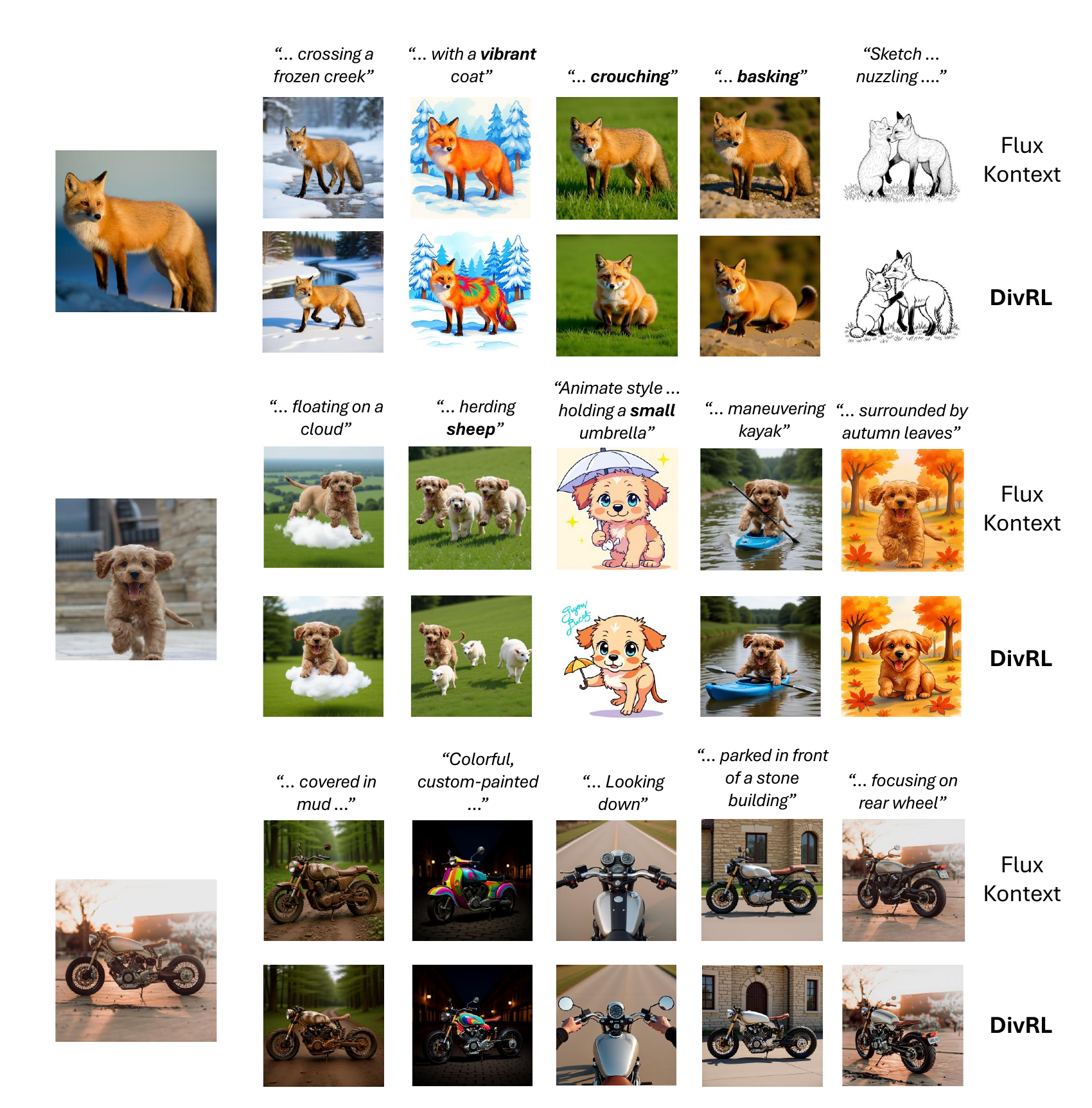}
\caption{Comparison between the original Flux-Kontext and DivRL under the multi-prompt setting. }
\label{fig:multi_prompt}
\end{figure}

In Figure \ref{fig:multi_prompt}, we present visual results of images generated from a single reference image across diverse text prompts. These results demonstrate our method's ability to maintain identity across multiple contexts. This is particularly evident in the more challenging "motorcycle" example, where our approach faithfully preserves fine-grained shapes and components across varying viewpoints.

\subsection{Multi-seed comparison}

\begin{figure}[t!]
\centering
\includegraphics[trim=0cm 3cm 0cm 0cm, width=0.9\textwidth]{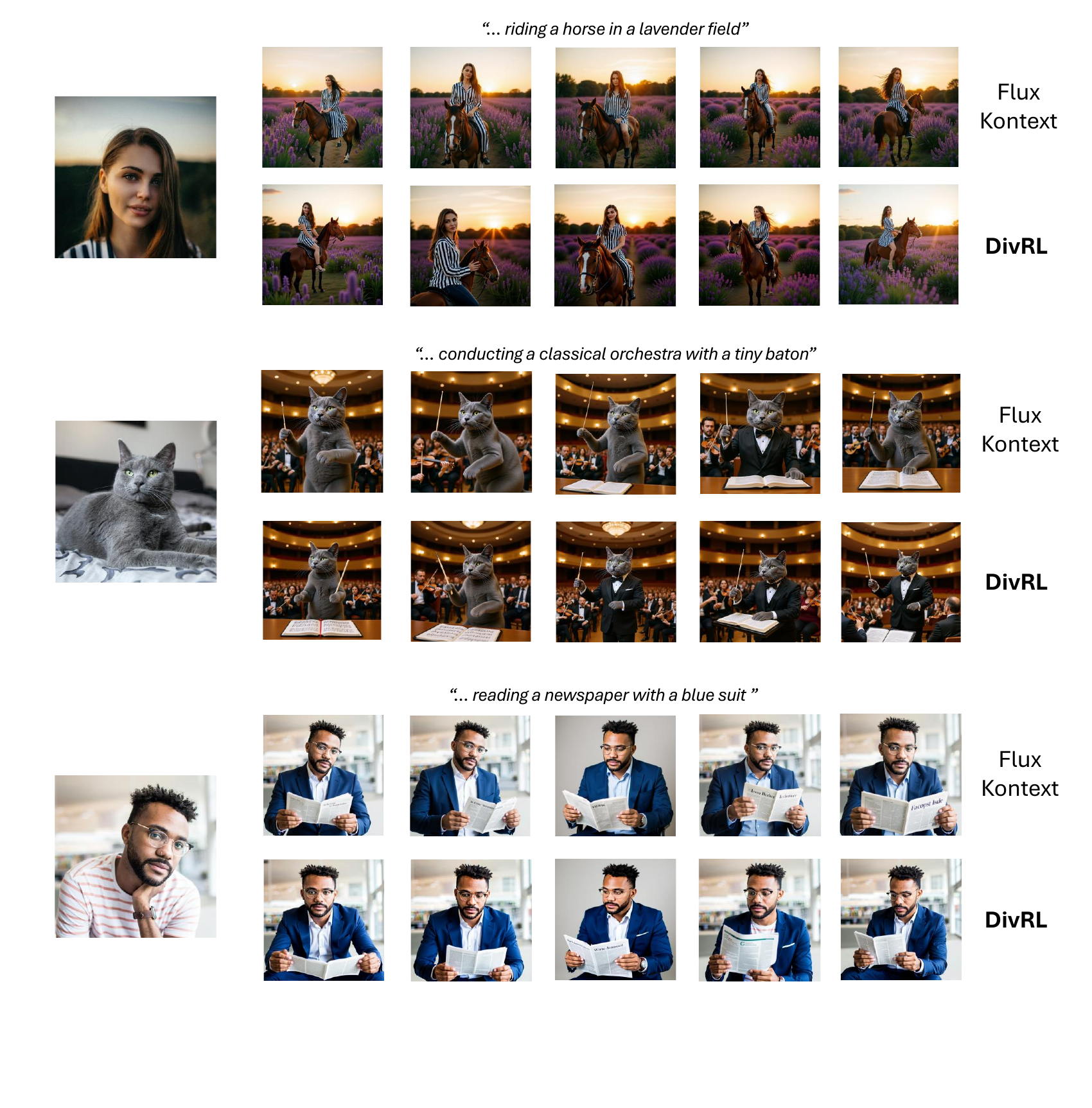}
\caption{Comparison between the original Flux-Kontext and DivRL under the multi-seed setting. }
\label{fig:multi_seed}
\end{figure}

In Figure \ref{fig:multi_seed}, we present visual results of image generated from a single reference image and a single text prompt under different random seeds. These results demonstrate our method's ability to produce diverse subject structures and viewpoints. Notably, in the first example of Figure \ref{fig:multi_seed}, the images generated by Flux-Kontext exhibit identity drift across different seeds, whereas our approach maintains consistent subject identity.

\subsection{Comparison with more baselines}

\begin{figure}[t!]
\centering
\includegraphics[trim=0cm 5cm 0cm 0cm, width=\textwidth]{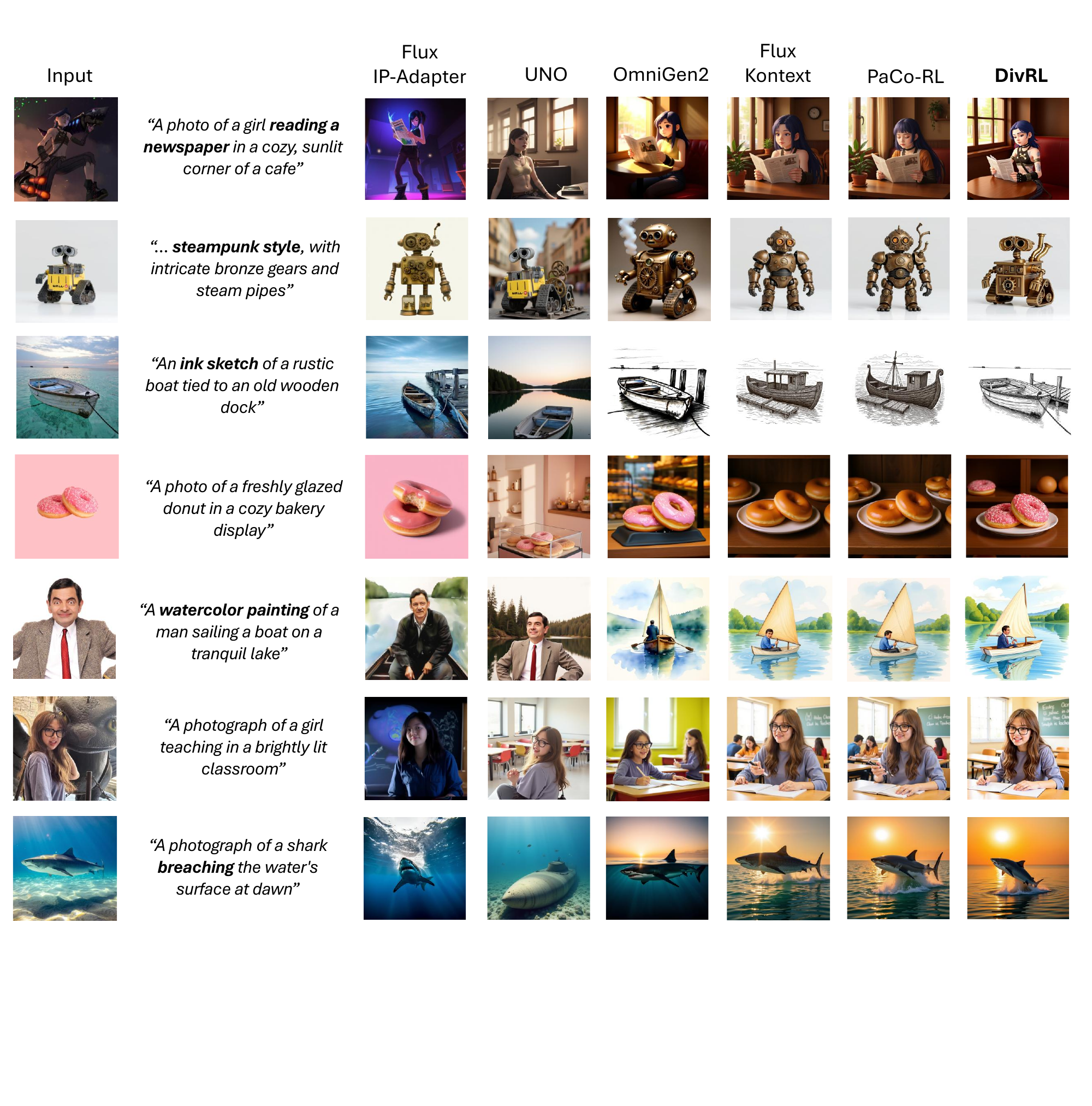}
\caption{Comparison between multiple baseline models and our method DivRL. }
\label{fig:quali_comp_supp}
\end{figure}

We provide an additional visual comparison between our method and other baseline models in Figure \ref{fig:quali_comp_supp}. These results demonstrate that our approach achieves superior subject identity consistency while simultaneously adhering to text prompts and maintaining high visual quality.

\section{Scope and failure cases}
\subsection{Scope analysis}
\begin{figure}[h]
\centering
\includegraphics[width=0.9\linewidth]{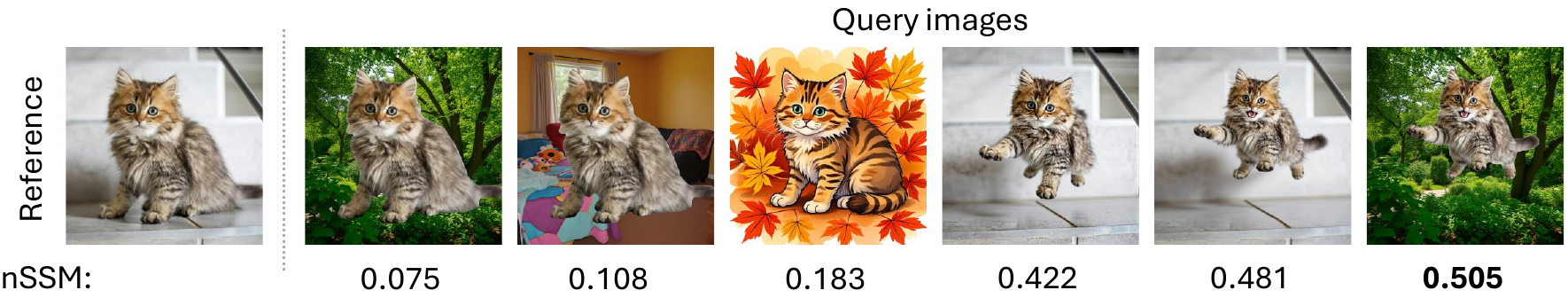}
\caption{How nSSM responds to semantic/structural changes. }
\label{fig:nssm_scope}
\end{figure}

We provide a qualitative analysis on how nSSM responds to semantic/structural changes in Fig.\ref{fig:nssm_scope}. When the semantics of the image are changing while the main foreground object remains the same, the nSSM score observes a medium increase. In contrast, when the structure (pose) of the main foreground object varies while the overall semantics say the same, the nSSM score has a sharper increase. nSSM score peaks when both structural and semantic changes exist. Therefore, despite being able to respond to both semantic and structural changes, nSSM is more sensitive to structural changes compared to the semantic ones, therefore encouraging more structural diversity in the RL finetuning stage.

\subsection{Failure cases}
\begin{figure}[h]
\centering
\includegraphics[trim=2cm 2cm 2cm 0cm, width=0.6\textwidth]{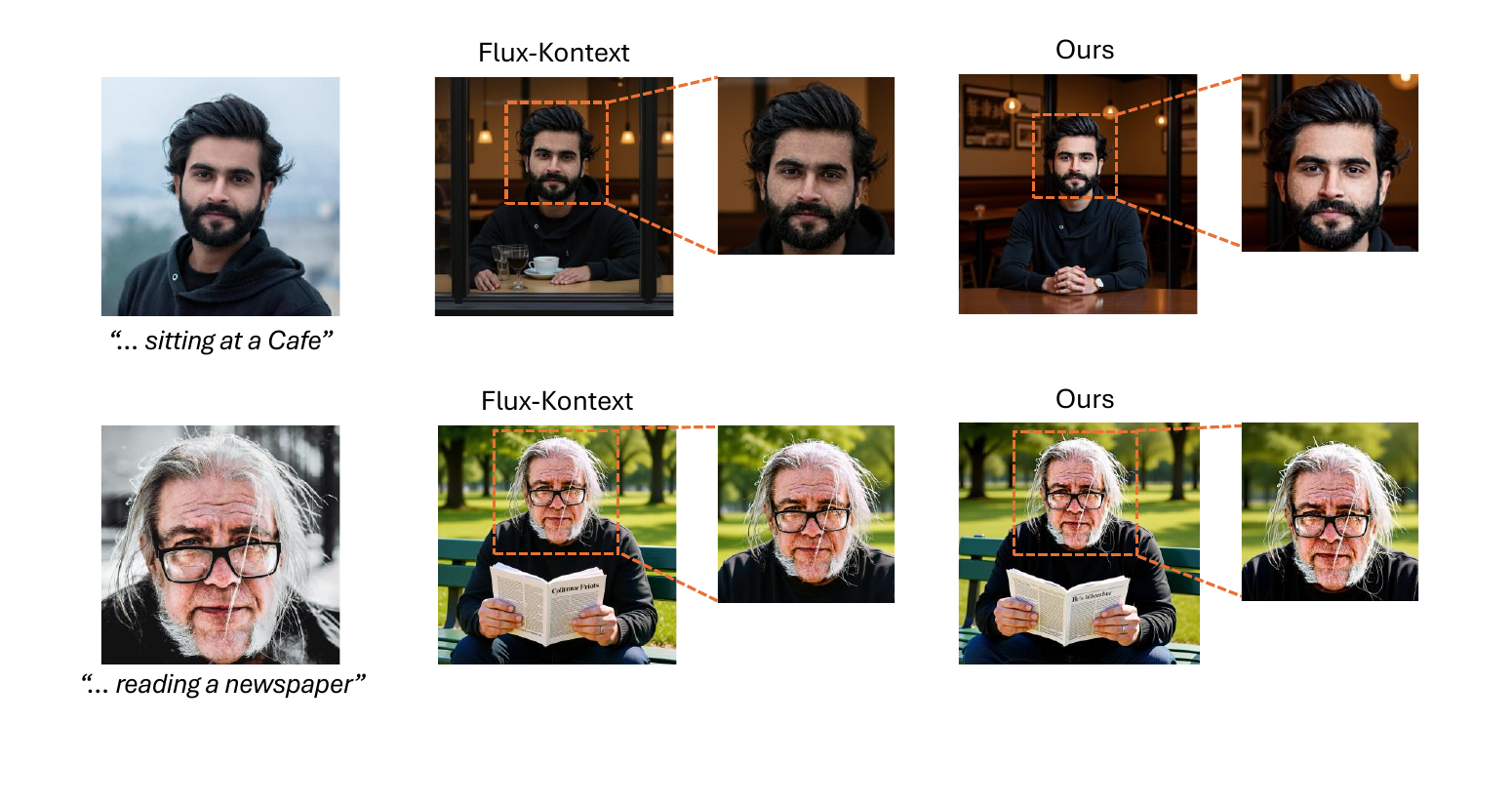}
\caption{Failure case: both the original Flux-Kontext model and our method may produce high-frequency noises of the identity region on the generated images. }
\label{fig:failure_noise}
\end{figure}

\begin{figure}[htbp]
\centering
\includegraphics[trim=0cm 10cm 0cm 0cm, width=0.9\textwidth]{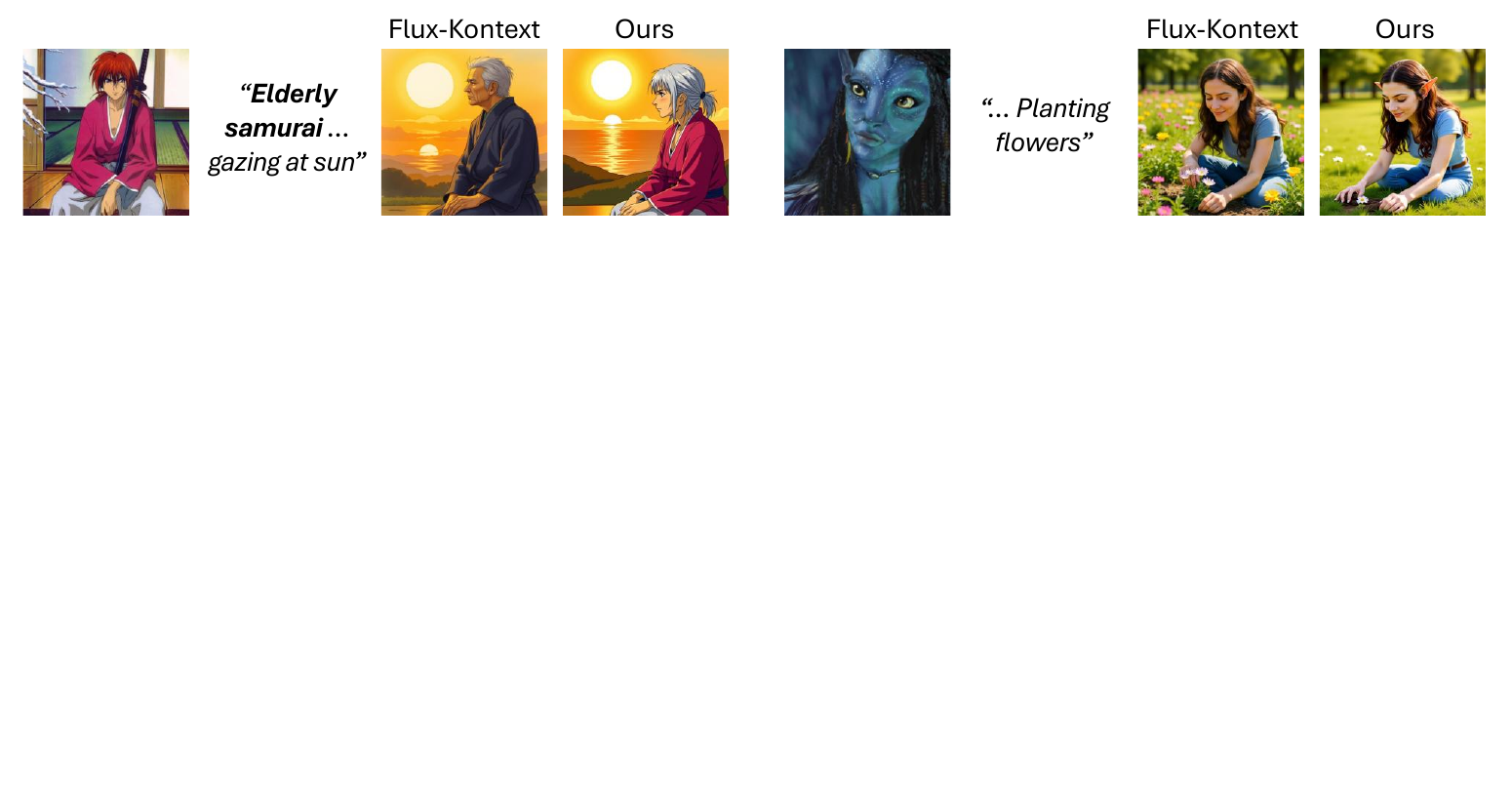}
\caption{Failure case: our method occasionally exhibits a trade-off between prompt adherence and identity consistency. }
\label{fig:failure_prompt}
\end{figure}

We observe that our method occasionally produces high-frequency artifacts within the subject's identity region, which is a phenomenon inherited from the base Flux-Kontext model as it is shown in Figure \ref{fig:failure_noise}. We hypothesize that this stems from the conditioning mechanism, which provides a strong local signal and emphasizes fine-grained visual features of the subject. During denoising, the model may over-amplify these signals, leading to high-frequency artifacts in the identity region. While our $24 \times 24$ spatial bottleneck serves as a spectral low-pass filter to mitigate the amplification of these artifacts during RL, we leave further architectural refinements for pixel-level visual fidelity to future work.

Additionally, despite its robustness, our method occasionally struggles to achieve an optimal balance between prompt adherence and identity consistency (Figure \ref{fig:failure_prompt}). In some instances, the model may prioritize the textual context at the expense of the unique features of the subject, leading to a partial loss of identity preservation; In some other cases, the model may prioritize the identity preservation at the expense of faithful prompt alignment.
\end{document}